\documentclass[12pt,A4]{article}
\usepackage[margin=1in,footskip=0.25in]{geometry}

\usepackage{bm,graphicx,color} 
\usepackage{amsmath,amsfonts}
\usepackage{xcolor}
\usepackage[linesnumbered,ruled,vlined]{algorithm2e}
\usepackage{hyperref}
\usepackage[numbers]{natbib}
\usepackage[flushleft]{threeparttable}
\usepackage{subfig}
\usepackage{graphicx}
\usepackage[toc,page]{appendix}

\SetKwInput{KwInput}{Input}                
\SetKwInput{KwOutput}{Output}              

\def\correspondingauthor{\footnote{Corresponding author: Jinwen.Qiu@wellsfargo.com}}

\begin{document}	
\title{Linear Iterative Feature Embedding: An Ensemble Framework for Interpretable Model}
\author{Agus Sudjianto, Jinwen Qiu \correspondingauthor{}, Miaoqi Li, Jie Chen
		}
\date{\normalsize Wells Fargo}	
\maketitle

\begin{abstract} 
A new ensemble framework for interpretable model called Linear Iterative Feature Embedding (LIFE) has been developed to achieve high prediction accuracy, easy interpretation and efficient computation simultaneously. The LIFE algorithm is able to fit a wide single-hidden-layer neural network (NN) accurately with three steps: defining the subsets of a dataset by the linear projections of neural nodes, creating the features from multiple narrow single-hidden-layer NNs trained on the different subsets of the data, combining the features with a linear model. The theoretical rationale behind LIFE is also provided by the connection to the loss ambiguity decomposition of stack ensemble methods. Both simulation and empirical experiments confirm that LIFE consistently outperforms directly trained single-hidden-layer NNs and also outperforms many other benchmark models, including multi-layers Feed Forward Neural Network (FFNN), Xgboost, and Random Forest (RF) in many experiments.  As a wide single-hidden-layer NN, LIFE is intrinsically interpretable. Meanwhile, both variable importance and global main and interaction effects can be easily created and visualized.  In addition, the parallel nature of the base learner building makes LIFE computationally efficient by leveraging parallel computing. 

\end{abstract}

\section{Introduction}
Ensemble methods have proved successful in the majority of machine learning competitions, as they integrate multiple machine learning algorithms into one predictive model. In particular, there are three main types of ensemble methods, including bootstrap aggregating (bagging), boosting, and stacking. Bagging and stacking learns base learners independently and aggregates them following a deterministic averaging process, while boosting learns sequentially in a very adaptive way and combines learners using a pre-specified strategy. The main goal of bagging is to arrive at an ensemble method with less variance than its base learners, whereas boosting mainly try to produce a strong model that is less biased than their base learners.  Stack ensemble aims at reduce both variance and bias by combining diversified base learners.  Compared with a single model, all of these ensemble methods can significantly improve predictive performance either by bias or variance reduction.

However, the final model produced by ensemble method is still regarded as a black box model, since the combination of base learners leads to a complicated model structure and makes inner decision-making process not transparent for human beings. The ability to explain the rationale behind one’s decisions to others is an important aspect of human intelligence in either social interaction or educational context. The interpretability of the results to enable business owners or regulators to better understand risk management decision processes and compel companies to meet regulatory requirements. Some commonly used interpretable models, such as linear model and general additive model cannot compete with ensemble models including Xgboost \cite{chen2015xgboost}  and Random Forest \cite{breiman2001random} since its simple structure cannot capture complicated dynamic data patterns.

In recent decades, some research works focus on enhancing interpretability by developing tools to ``open up the black box". There are, broadly speaking, three inter-related model–based areas of research: a) global diagnostics (Sobol \& Kucherenko (2009) \cite{kucherenko2009derivative}, Kucherenko (2010)  \cite{kucherenko2010new}); b) local diagnostics (Sundararajan et al. (2017) \cite{sundararajan2017axiomatic}, Ancona et al. (2018) \cite{ancona2017towards}); and c) development of approximate or surrogate models that may be easier to understand and explain. However, (Rudin and Cynthia (2019) \cite{rudin2019stop} ) suggests to avoid using explainable black-box models in high-stakes decisions since they are sometimes problematic with several reasons such as unreliable presentation or lack of details of what the original model delivers. Therefore, other researchers made efforts to build inherently interpretable model, such as explainable neural network (xNN) proposed by (Vaughan et al. (2018) \cite{vaughan2018explainable}), adaptive explainable neural networks (AxNNs) by (Chen et al. (2020) \cite{chen2020adaptive}), and explainable neural networks with constraint (Yang et al. (2020) \cite{yang2020enhancing}).  Following this research direction, we try to build an inherently interpretable with a predictive performance as strong as some black box models or an even better performance.

In this paper, our LIFE algorithm fulfills three main goals: competitive predictive performance, boosted computation efficiency, and interpretable model.  We know that the single-hidden-layer NN has universal approximation property in theory and is easy to be interpreted due to simple architecture.  However, we need to resort to a wide single-hidden-layer NN with a large number of neural nodes to obtain a strong predictive performance, which is numerically difficult to estimate in practice. Therefore, by leveraging the ensemble method and a simple structure of a single-hidden-layer NN, we developed an innovative and flexible framework called LIFE to train wide single-hidden-layer NNs, which can achieve both high accuracy and easy interpretability in both regression and classification settings.

In this algorithm, we first use a special hierarchical structure of multiple single layer NNs to perform data sampling based on the linear projection of neurons, and then train multiple narrow single-hidden-layer NNs with ReLU activation as base learners on different subsets of the dataset; finally, we aggregate neural nodes as features from multiple base learners into a wide single layer NN and do a join estimation via a linear model. Compared with the traditional training strategy, LIFE effectively avoids directly training a wide single layer NN by extracting features from multiple narrow single-hidden-layer NNs trained on different subsets of dataset instead. In this way, we can introduce diversity among the base learners, which tends to decrease the total uncertainty after ensemble and, thus, yields better results empirically. To achieve good diversity among base learners, we build a hierarchical structure of multiple single neural networks, and leverage the linear projections of the neurons to define the subset of sampling. In addition, the algorithm combines the features defined by neurons from the single layer NN base learner, for which we called neurons flattening, and this can further improve results compared with traditional ensemble methods. This technique also allows LIFE to take advantage of parallel computing to improve computational efficiency through training multiple narrow single-hidden-layer NNs simultaneously. 

Extensive analyses on both simulated data and public real data verify its effectiveness in both predictive and computational performance. Furthermore, we provide theoretical foundation for LIFE and prove the importance of the diversity of base learners by exploring the relationship with two-stage ensemble stacking and the ambiguity loss decomposition for two-stage ensemble stacking. The final single-hidden-layer NN architecture obtained from LIFE allows to visualize the neural network weights and bias and understand the input and output relationship easily. In particular, a single-hidden-layer NN with rectified linear unit (ReLU) activation function \cite{vaughan2018explainable} is equivalent to an additive index model with linear splines on linear projections. Moreover, it can be considered a local linear model, where all predictors are easily visualized by a parallel coordinates plot \cite{andrienko2001constructing}. The main and interaction effects can also be identified by aggregating local linear model coefficients. 

In general, our main contributions are summarized below:
\begin{enumerate}
\item LIFE is an innovative and flexible framework for ensemble methods, which allows different kinds of variants.
\item LIFE can achieve a better predictive performance than traditional single-hidden-layer NN training methods, as demonstrated by the theoretical background and empirical experiments.
\item LIFE can still keep model interpretable, and a new interpretation tool is introduced to detect main and interaction effects.
\item  LIFE can improve the computation efficiency via easy parallelization, rendering wide single layer NN training faster.
\item  An theoretical foundation for LIFE based on ambiguity loss decomposition and diversity of base learners is provided. 
\end{enumerate}

The rest of the paper is organized as follows: In Section \ref{sec:meth}, we introduce LIFE algorithm and theoretical rationale behind LIFE through loss decomposition. Extensive experiments on simulated and real data are conducted in Section \ref{sec:ee} to test the performance of LIFE under various conditions in comparison with other benchmark algorithms. In Section \ref{sec:int}, we explored interpretation of model such as main or interaction effect detection.  In section \ref{dis}, we discuss the model pruning and the extension of LIFE algorithm in more depth. In Section \ref{sec:con}, we provide our conclusions. In section 6, we discuss the model pruning and the extension of LIFE algorithm in more depth.

\section{Methodology}\label{sec:meth}
In this section, we introduce a new proposed ensemble method called LIFE. The LIFE algorithm is mainly used to train a wide single-hidden-layer NNs in an iterative way.  First, the general framework for LIFE including three steps is introduced in Section \ref{sec:gf}. Then, details of LIFE algorithms is provided in Section \ref{sec:LA}.  Finally the theoretical foundation is discussed in Section \ref{sec:TF}. 

\subsection{General framework for LIFE} \label{sec:gf}
\begin{figure}[h]
\center
\includegraphics[width=16cm,height=4cm]{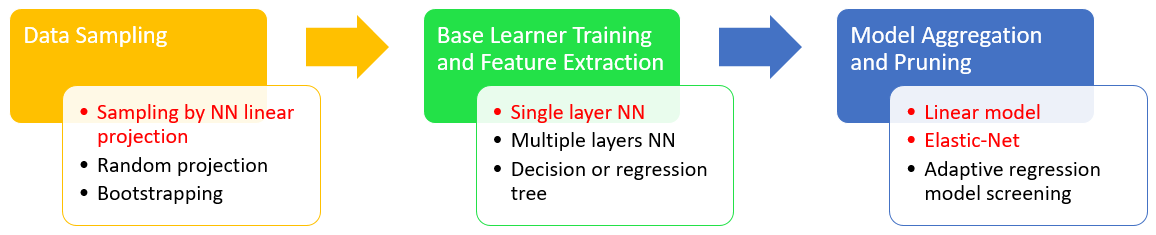}
\caption{It is the general framework of LIFE algorithm and options colored in red in each step are used in the paper.}
\label{fig:fw}
\end{figure}

As shown in the Figure  \ref{fig:fw}, the LIFE algorithm consists of three steps: data sampling, base learner training and feature extraction, model aggregation and pruning.  Figure \ref{fig:fw} presents several options in the white box for each step, which allow for various combinations of those steps to achieve pre-specified goals.

\begin{enumerate}
\item  The first step consists in defining subsets of data via active functions of NN neurons, which are obtained by training multiple single-hidden-layer NNs in a hierarchal structure as illustrated in Figure  \ref{fig:tree}.   The diversified base learners can be generated by training based on these subsets for data sets.  More theoretical explanation on ensemble with diversified base learners will be provided in Section \ref{sec:TF}.   There are other alternative ways to generate sampling for base learner training, e.g., bootstrapping in the traditional method or data splitting via random projection. Through leveraging linear projections from trained NNs, our sampling method in the supervised setting can more effectively generate the diversity among base learners, which is demonstrated by empirical experiments in Section \ref{sec:dss}.

\item The second step consists in training base learners on different subsets of data sampled during the first step.  Various options can be considered, e.g., single layer NN, multiple layer NN, and regression or decision tree, etc., In this paper, the single-hidden-layer NN is used as the base learner given its interpretability. After estimating multiple NN base learners, all activation functions of the neurons from NN base learners are extracted as new features.

\item The third step consists in combining all new extracted features from different base learners in the second step to construct the final predictive model. In addition, new features can be pruned through regularization or other methods to generate a more parsimonious model.   We use linear model and elastic–net in our LIFE algorithm which is simple and straightforward in this paper, but some other more complicated model aggregating methods e.g., adaptive regression model screen, and pruning methods, e.g., base learner selection algorithm \ref{algo:bls}, will be discussed in Section \ref{sec:bls}.
\end{enumerate}

\subsection{LIFE Algorithm}\label{sec:LA}
LIFE algorithm is an iterative process with multiple single layer NN base learners trained in each iteration.  Assume there are $J$ iterations. The first $J-1$ iterations is used to define the data sampling through a hierarchical structure, and the last $J$ iteration is used to build the features from single layer NN base learners. Moreover, let $[K_1,\cdots, K_J]$ denote the collection of the number of hidden neurons for single-hidden-layer NNs in all iterations, where $K_j$ is the number of hidden neural nodes for single-hidden-layer NNs which are trained in the $j^{(th)}$ iteration. Figure \ref{fig:tree} below gives an illustration to LIFE framework, with $[K_1, K_2, K_3] = [3,3,2]$, where $\hat{b}_k^{(j)}$'s and $\hat{w}_k^{(j)}$'s represent biases and weights from single-hidden-layer NNs respectively, $\hat{\beta}_k$'s are coefficients used to linearly combine all new neurons in the final step, and the $cp$ is the cutoff point to define subsets by controlling subset size.

\begin{figure}[h]
\center
\includegraphics[width=18cm,height=11cm]{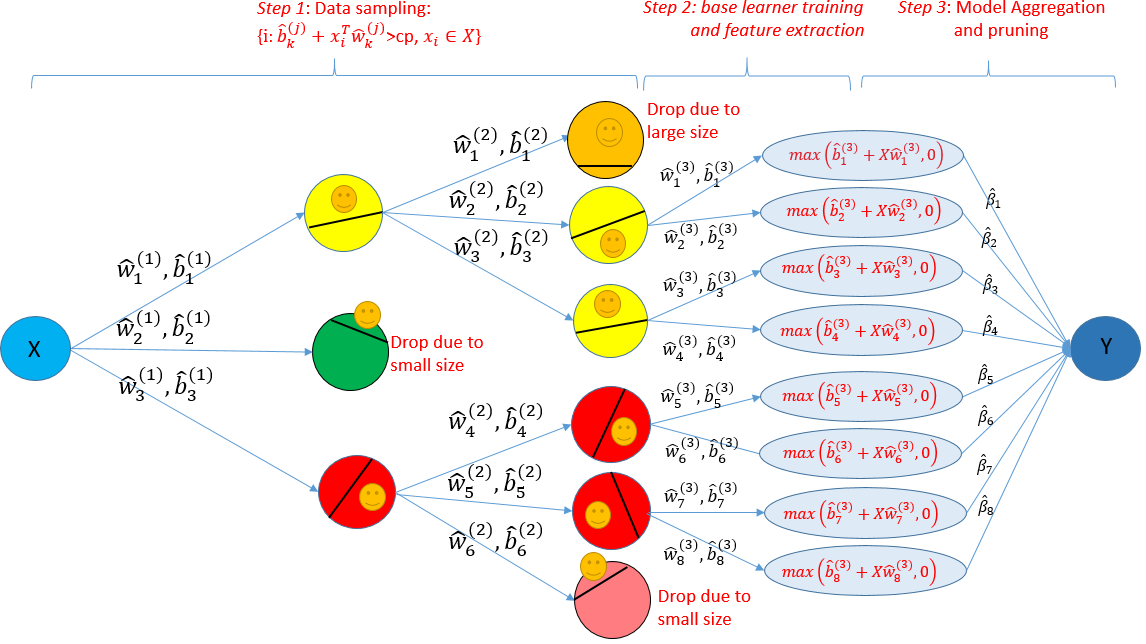}
\caption{An illustration of LIFE framework with $[K_1,K_2,K_3] = [3,3,2]$. Regions with a smiling emoji face indicate that the subsets satisfying the sampling conditions and are used for NN models training in the next iteration.}
\label{fig:tree}
\end{figure}

\begin{figure}[h]
\center
\includegraphics[width=17.5cm,height=4cm]{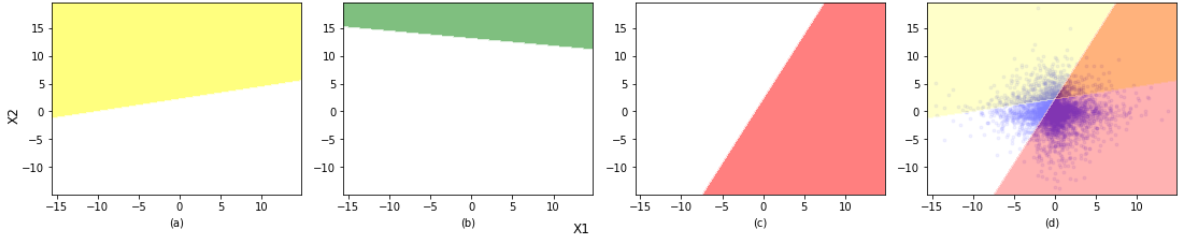}
\caption{That is data partition by NN linear projection after the first iteration and observations in the colored area from (a), (b), (c) will be selected in the each subset. The (d) shows two kept data partitions in one plot.}
\label{fig:part1}
\end{figure}

The LIFE framework illustration in Figure \ref{fig:tree}  can be separated into three steps as displayed in Figure 1. The two iterations in the first step are used to perform data sampling, in which we first fit a single-hidden-layer NN with three hidden neurons, then define subsets by $b_k^{(1)}+ x_i^Tw_k^{(1)}>cp$, where $k=1,2,3;i=1,\cdots,N$,  given estimated bias and weight in each neuron.  Further, the entire neuron is dropped and no longer be used for next iteration if the defined subset in this neuron is either too large or too small based on pre-specified criteria. For example, the green neuron in the middle is dropped due to small size of subset. If the subset size is close to the full data size, the sample is almost identical to original training data, which is not beneficial for generating diversified samples.  On the other hand, if the subset size is too small, the sample is not representative of the original data, and the base learner built on this sample does not have good performance on the entire training data. We will show in Section 2.3, the diversity and accuracy of the base learners are the two key elements for the final ensemble performance.  As the result of the first iteration in the first step, we leverage NN linear hyperplane in the neurons to perform data partition as shown in Figure \ref{fig:part1}, of which the idea is similar to oblique trees  \cite{heath1993induction}. Plots (a), (b) and (c) in Figure \ref{fig:part1} have shown how data are partitioned in different ways for these three neurons based on $ w_k^{(1)} $'s and $ b_k^{(1)} $'s, $k = 1,2,3$, obtained from the first iteration. The non-white regions (yellow, green or red) indicate the subsets, where all observation satisfy $b_k^{(j)}+ x_i^Tw_k^{(j)}>cp$, and will be used to fit single-hidden-layer NNs in the second iteration. Notice that the non-white region in plot (b) is so small, which corresponds to the dropped green neuron in Figure \ref{fig:tree}. Plot (d) in Figure \ref{fig:part1} shows the combination of plot (a) and (b). As we can see, the two subsets are overlapped, which is different from the subsets defined in the traditional regression or decision tree structure, which uses the exclusive partition.

\begin{figure}[h]
\center
\includegraphics[width=17.5cm,height=8cm]{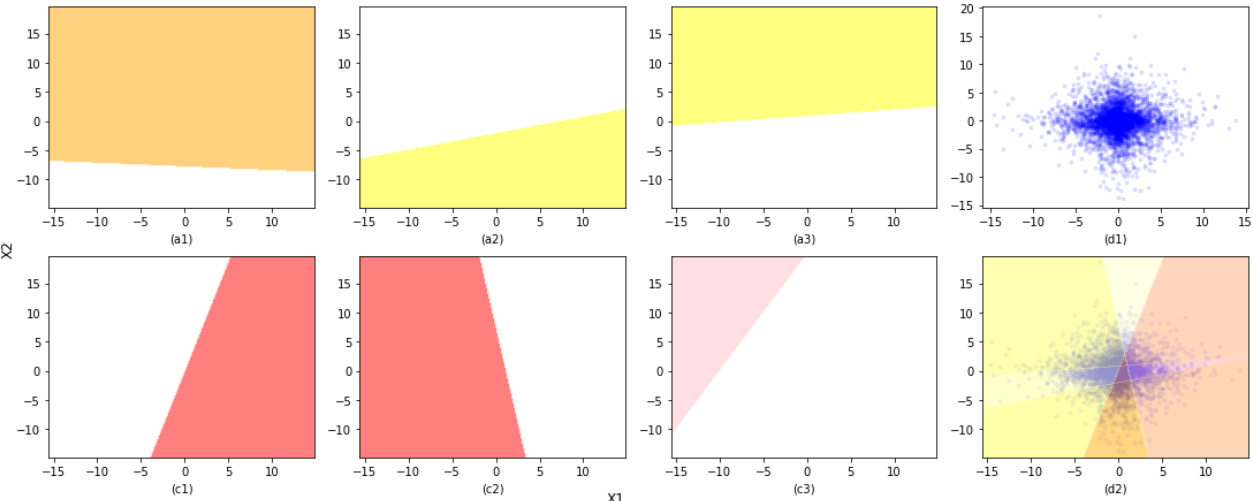}
\caption{It is data partition by NN linear projection after the second iteration and observations in the colored area from (a1-a3), (c1-c3) will be selected in the each subset. The (d1) displays all training data points in green, and the (d2) shows four kept data partitions in one plot.}
\label{fig:part2}
\end{figure}

In the second iteration of the first step, single-hidden-layer NNs with three hidden neurons are fitted independently using the subsets defined from the first iteration. After that, the entire training set is evaluated for data sampling through NN linear projection $b_k^{(2)}+ x_i^Tw_k^{(2)}$, where $k=1,\cdots,6; i=1,\cdots,N,$ and new subsets are defined satisfying $b_k^{(2)}+ x_i^Tw_k^{(2)}>cp$. Note that, the new subsets are defined on the entire training data, not the samples from the previous iteration, which is also very different from traditional regression or decision tree structure. Again, neurons with too small or too large subset are dropped forever, as shown in the first node (in brown) and the last node (in pink) of the second hidden layer from Figure \ref{fig:tree}, corresponding to (a1) and (c3) in Figure \ref{fig:part2}. In addition, Figure \ref{fig:part2} shows data partition obtained from the second iteration, where plots (a1), (a2) and (a3) represent the partitions generated from NN trained on the subset of the first node (yellow) in the first iteration, while plots (c1), (c2) and (c3) correspond to the third node (red) in the first iteration. Plot (d2) shows that each training data point as least belongs to one of the six subsets in (a1-a3) and (c1-c2). We do expect all or most data points are covered by different subsets. The reason is that the neurons with different active regions in NN are representing different features and patterns from the data, LIFE trains the same type of base learner model on part of the dataset but evaluates it on the entire dataset. This leads to small errors appearing in the region of sampled data and large errors outside region. Therefore, data sampling with these active regions can effectively define subsets for generating more diverse representation of data and producing less correlated prediction errors.  The sampling in this supervised manner is better than sampling in a random way and this is further discussed in Section \ref{dis}. In practice, the first step can be reduced to one iteration or have more than two iterations, displayed in Algorithm \ref{algo:LIFE}.

\begin{algorithm}
\DontPrintSemicolon
  \KwInput{$[K_1,\cdots, K_J]$, upper bound: $u$, lower bound: $l$, cutoff point: $cp$ }
  \KwOutput{$\hat{\beta}_k,\hat{b}_k^{(J)}, \hat{w}_k^{(J)}$, $k=1,\cdots, m_{j}$, where $m_{j}$ is the total number of neural nodes left in the $j^{th}$ iteration.}
  \KwData{Training set $\{x_i,y_i\}_{i=1,\cdots,N}$}
      \For{$j=1,\cdots,J$}    
        { 
           \If{$j=1$}
             {
            1 .Train a single-hidden-layer NN regressor or classifier with the number of hidden neurons equal to $K_j$ on training set $x,y$ by specified optimizer. \\
            2. Collect estimated parameters $\hat{b}_k^{(j)}, \hat{w}_k^{(j)}$, where $k=1,\cdots, K_j$. \\
            3. $m_j=K_j$
            }
     \Else
    {
    	  \For{$k=1,\cdots, m_{j-1}$}
     { 1. Initialization: $M=0$ , where $M$ records the number of NNs trained in the current iteration.\\
       2. Sample observations from training set  $\{x_i,y_i\}_{i=1,\cdots,n}$ based on $\hat{b}_k^{(j-1)}+ x_i^T\hat{w}_k^{(j-1)}>cp$, where $x_i\subseteq \{x_i\}_{i=1,\cdots,n}$.\\ 
       3. Record size of sampled observations as $s$.\\
       4. \If{$l<s/N<u$ }    
        {1. Train a single-hidden-layer NN regressor or classifier with the number of hidden neurons equal to $K_j$ on sampled subset of training set $x,y$ by specified optimizer. \\
         2. Collect estimated parameters $\hat{b}_q^{(j)}, \hat{w}_q^{(j)}$, where $q=1,\cdots, K_j$.\\
        3.  $M=M+1$
}
        
         4.  \Else
    {Skip this step.}
    5. $m_j=M\times K_j$  
   }
    }

        }  
From  $M$ trained NNs, collect  $\hat{b}_k^{(J)}, \hat{w}_k^{(J)}$, where $k=1,\cdots, m_{J}$.  \\
Calculate values of each neural nodes $\sigma(\hat{b}_k^{(J)}+ x^T\hat{w}_k^{(J)})$, which are called new features. \\
Fit a linear regression or logistic model with or without regularization using new features as input to estimate $\hat{\beta}_k$, where $k=1,\cdots, m_{J}$. 
\caption{LIFE}
\label{algo:LIFE}
\end{algorithm}

In the second step, four single-hidden-layer NNs, each with two hidden neurons are trained as base learners independently on the subsets defined at the end of the first step, and then the features are generated by the ReLU activation functions.  Please note that the features are evaluated on the entire training dataset.   
In the third step, we combine all these new features together.Since all features are obtained based on the entire training set, which technically forms a design matrix with dimension $N \times m_J$, where $N$ is the size of training set and $m_J$ is the total number of extracted features from step 2 with $J=8$ shown in Figure \ref{fig:tree}. Then a linear model is fitted on these features and $\beta_i, i = 1, \cdots, m_J,$ is the coefficient of each feature. In the default setting, the linear regression or logistic model is applied to combine neural nodes extracted from different base learners and make a final prediction. However, there may be too many features and some of them can even be highly correlated, leading to overfitting problem. Therefore, we need to prune some redundant neural nodes through adding regularization or removing some base learners. Both methods can not only prevent overfitting, but also produce a more parsimonious NN model with fewer nodes that is beneficial for interpretation. As a regularized regression method that linearly combines the $L_1$ and $L_2$ penalties, the elastic net is one of the options, and it is used in the paper for the third step of model aggregation and pruning. LASSO and ridge regression are treated as special cases of elastic net.

By combining multiple relatively narrow but diversified single-hidden-layer NNs with $K_J$  hidden neurons for each, the LIFE algorithm finally constructs a wide single-hidden-layer NN with $m_J$ hidden neurons, which is $8$ in Figure ref{fig:tree}. This trained process is completely different from the traditional methods, with which a NN is optimized as a whole by stochastic gradient based optimizers. In most cases, It is numerically difficult and computational expensive to train a single layer NN and achieve good performance. LIFE can help overcome this difficulty and achieve decent performance. In addition, LIFE can leverage parallel computation to significantly reduce the training time. 
In the end, we provide the detailed pseudo-code of LIFE in $J$ iterations,  described in Algorithm \ref{algo:LIFE}, where $m_j$ indicates the number of remaining neural nodes after the $j^{th}$ iteration in the first step, where $j=1,\cdots,J$. Both $u$ and $l$  are maximal and minimal proportions of training set size, which provides the upper and lower bound for subset size, respectively. The neuron will be dropped if the proportion of subset size is beyond the range.


\subsection{Theoretical Foundation}\label{sec:TF}
The ensemble method, such as stacking \cite{wolpert1992stacked}, bagging \cite{breiman2001random}, boosting \cite{chen2016xgboost} or Bayesian model averaging \cite{qiu2018multivariate, qiu2020multivariate}  is composed of a multiple independently trained regressors or classifiers whose predictions are combined to make final predictions. Empirically, ensembles tend to yield better results than a single model when there is a significant diversity among the models \cite{kuncheva2003measures}.  For the past few decades, many studies have been focusing on accuracy and diversity of ensemble methods in either regression \cite{brown2005managing} or classification case \cite{aksela2006using, gacquer2009effectiveness, butler2018effectiveness}.  (Krogh and Vedelsby  (1994) \cite{krogh1994neural}) proposed ambiguity decomposition and a computable approach to minimize the quadratic error of the ensemble estimator, while (Ueda and Nakano (1996) \cite{ueda1996generalization}) derived a general expression of bias-variance-covariance decomposition. (Brown et al. (2005) \cite{brown2005diversity}) and (Hansen (2000) \cite{hansen2000combining}) investigated the connections between ambiguity decomposition and bias-variance-covariance, and have shown they are identical. Based on ambiguity decomposition, we establish the theoretical foundation for LIFE and extend loss decomposition for both regression or classification setting, which will be discussed in Section \ref{Connection} and \ref{LossDecomp}.

\subsubsection{Connection to Stacking} \label{Connection}
Stacking is a type of ensemble method, by which a final model is trained from the combined predictions of another models. In stacking, the predictions from different machine learning models are used as new inputs and are combined to generate a new set of predictions.
Those predictions can be used on additional layers, or the process can stop here with a final result. One important assumption behind stacking is that different base learners can produce weakly correlated prediction errors that are complementary. If we use weighted averages, we might believe that some of the base learner are better or more accurate and can be assigned higher weights. In the framework of stacking, an even better approach might be to estimate these weights more intelligently by using another layer of the learning algorithm, such as the linear model. 

\begin{table}[h]
\centering
    \begin{tabular}{ | l | p{5cm} | p{5cm} |}
    \hline
     Aspect & Stacking  & LIFE  \\ \hline
    Base Learner & different models or same model with different settings  &  same model with same setting trained on different subsets of data  \\  \hline
    Model averaging  & linear or nonlinear combination of prediction from base learners  & linear or nonlinear combination of features flatten from base learners \\
\hline
    \end{tabular}
\caption{Difference between LIFE and Stacking}
\label{tab:stack}
\end{table}
\begin{figure}[h]
     \centering
    \subfloat[][MIM]{\includegraphics[width=3.2in]{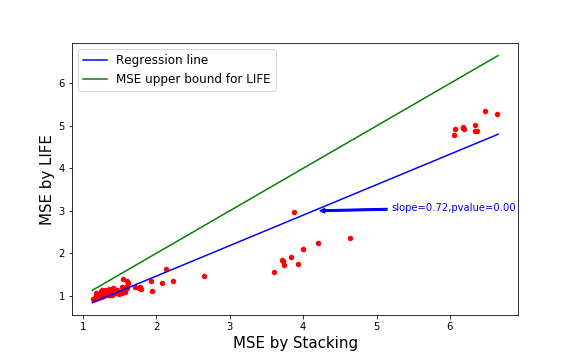}\label{fig:mim}}
     \subfloat[][California Housing]{\includegraphics[width=3.2in]{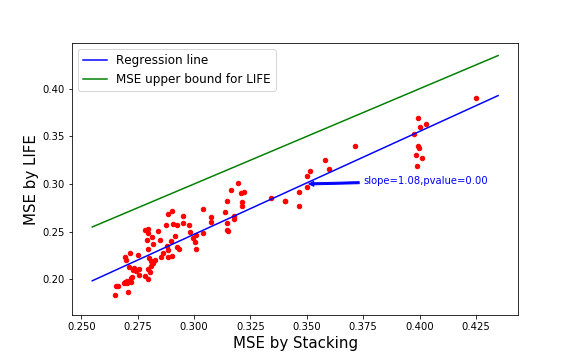}\label{fig:cal}}
     \caption{Relationship between LIFE and Stacking (Regression)}
     \label{fig:stack_life_reg}
\end{figure}
\begin{figure}[h]
     \centering
    \subfloat[][MIM]{\includegraphics[width=3.2in]{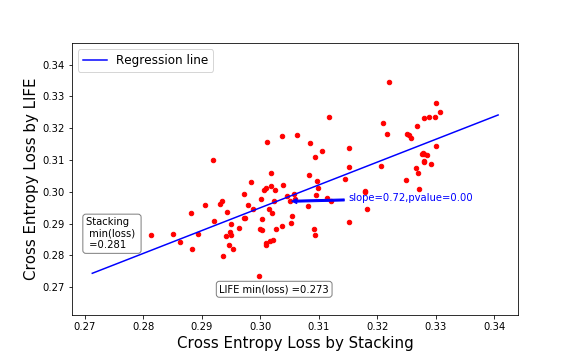}\label{fig:mim}}
     \subfloat[][Gamma Telescope]{\includegraphics[width=3.2in]{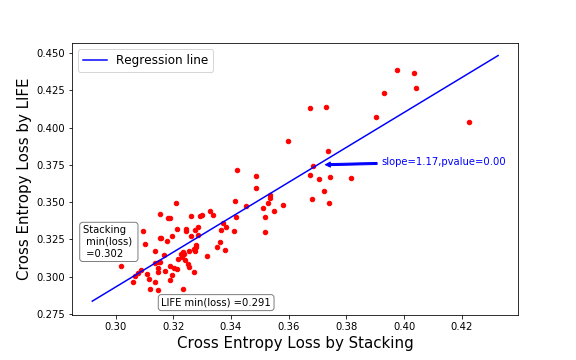}\label{fig:cal}}
     \caption{Relationship between LIFE and Stacking (Classification)}
     \label{fig:stack_life_cal}
\end{figure}

Some major differences between stacking method and LIFE algorithm as shown in Table \ref{tab:stack}. Without neural nodes flattening, LIFE is very similar to stacking.Despite the differences between the two methods, the minimization of loss function of LIFE is approximately equivalent to the  minimization of loss function for the two-stage stacking method, which firstly fits multiple base learners (NN trained on different subsets sampling by LIFE algorithm) and use predictions from base learners as input to train a model averaging model. Figure \ref{fig:stack_life_reg} and \ref{fig:stack_life_cal} indicate a strong linear relationship between LIFE and the stacking method with different hyper-parameter setup in terms of MSE loss or cross entropy loss in both regression and classification cases. This linear relationship has be verified by both simulated data (MIM) and real data (California Housing for regression and Gamma Telescope for classification). Due to the BLUE (Best linear unbiased prediction) property of OLS estimator in linear regression, the joint estimation of the coefficients of all the combined features in the three step makes LIFE always outperform two-stage stack ensemble method with the same setting.  This can be verified in Figure \ref{fig:stack_life_reg} that all the points are below the green diagonal line. For classification case, LIFE also performs better than stacking method with smaller minimum loss as shown in the white box of Figure \ref{fig:stack_life_cal}.

\subsubsection{Loss Function Decomposition} \label{LossDecomp}
(Krogh and Vedelsby  (1994) \cite{krogh1994neural}) proposed ambiguity decomposition for quadratic error of the ensemble estimator which is the sum of the quadratic loss of individual base learners and the ambiguity measure for diversity.  We extend the ambiguity decomposition to both mean square error and cross entropy error via Taylor expansion, where the two loss functions corresponds to regression and classification, respectively.  
Let an ensemble model with M base learners be expressed as  $f_{ens}= \sum_{j=1}^M \beta_jf^{(j)}$, where $\sum_{j=1}^M\beta_j=1$ and $\beta_j\geq0$.

For any loss function that is twice differentiable, we can expand the loss function of $j^{th}$ base learner around output of an ensemble model based on Taylor's theorem with Peano form of the remainder as follows:
\begin{equation} \label{eq:taylor0}
\begin{gathered}
l(y,f^{(j)}) = l(y,f_{ens})+ l^{\prime}(y,f_{ens})(f^{(j)}-f_{ens}) + \frac{1}{2} l^{\prime\prime}(y,f^{(j)\star})(f^{(j)}-f_{ens})^2,
\end{gathered}
\end{equation}
where  the value of $f^{(j)\star}$  is between $f_{ens}$ and $f^{(j)}$.  Multiplying both sides of equation \eqref{eq:taylor0} by $w_j$ and taking a sum yield:
\begin{equation} \label{eq:taylor}
\begin{gathered}
 \sum_{j=1}^{M} \beta_jl(y,f^{(j)}) = \sum_{j=1}^{M} \beta_j l(y,f_{ens})+  \sum_{j=1}^{M} \beta_j l^{\prime}(y,f_{ens})(f^{(j)}-f_{ens}) \\
+ \frac{1}{2}  \sum_{j=1}^{M} \beta_jl^{\prime\prime}(y,f^{(j)\star})(f^{(j)}-f_{ens})^2.
\end{gathered}
\end{equation}
The second term on the right side of \eqref{eq:taylor} is expressed by:
\begin{equation} \label{eq:taylor2}
\begin{gathered}
  \sum_{j=1}^{M} \beta_j l^{\prime}(y,f_{ens})(f^{(j)}-f_{ens})= l^{\prime}(y,f_{ens})\{\sum_{j=1}^{M} \beta_j f^{(j)}-f_{ens}\sum_{j=1}^M\beta_j\}\\
     =l^{\prime}(y,f_{ens})\{f_{ens}-f_{ens}\}=0.
\end{gathered}
\end{equation}
Since this term is zero, the loss function $l(y,f_{ens})$ of the ensemble can be decomposed into:
\begin{equation} \label{eq:dec}
\begin{gathered}
l(y,f_{ens}) = \sum_{j=1}^{M} \beta_j l(y,f^{(j)}) - \frac{1}{2} \sum_{j=1}^{M}\beta_j l^{\prime\prime}(y,f^{(j)\star})(f_{ens}-f^{(j)})^2.
\end{gathered}
\end{equation}
In regression case, let $f_i=\sum_{j=1}^M \beta_j f_{i}^{(j)}$ be individual predicted value of an ensemble model for $i^{th}$ observation, where $M$ is the number of base learners, $f_{i}^{(j)}$ represents individual predicted value of $j^{th}$ base learner for $i^{th}$ observation and $\beta_j$ denotes regression coefficient for $j^{th}$  base learner. 
Basically,  the mean squared error (MSE) is commonly used loss function $l(y,f)=\frac{1}{N}\sum_i^N(y_i-f_i)^2$ for regression problems, where $N$ is the total number of observations in the entire dataset. Based on equation \eqref{eq:reg_dec}, MSE of an ensemble model can be written in terms of the ambiguity decomposition given $x_i, i=1,\cdots,N$:
\begin{equation} \label{eq:reg_dec}
\begin{gathered}
MSE = \frac{1}{N}\sum_{i=1}^N(y_i-f_i)^2= \frac{1}{N}\sum_{i=1}^N(y_i-\sum_{j=1}^M \beta_j f_{i}^{(j)})^2 \\
=\underbrace{\frac{1}{N}\sum_{i=1}^N\sum_{j=1}^M \beta_j(y_i- f_{i}^{(j)})^2}_\text{accuracy}-\underbrace{\frac{1}{N}\sum_{i=1}^N\sum_{j=1}^M \beta_j(f_i- f_{i}^{(j)})^2}_\text{diversity}.
\end{gathered}
\end{equation}
On the right-hand side of equation \eqref{eq:reg_dec}, the first term of this decomposition is to measure average prediction accuracy of base learners, while the second term is called ambiguity (hence the name of the decomposition) and can be easily interpreted in terms of diversity between individual base learners. Unlike the bias-variance-covariance decomposition, the ambiguity decomposition highlights a trade-off between the average accuracy of base learners, and their deviation from the ensemble output.

Regarding LIFE, the base learner is a single-hidden-layer neural network trained on a subset of all observations. A stronger base learner indicates a better performance of model, which is reflected by first term of equation \eqref{eq:reg_dec}. If the subset size is small, the base learner is also weak, which deteriorates performance. Thus, a lower bound is set up. The power of LIFE framework comes from second term diversity, which is due to data sampling during iterations. Creating different subsets through sampling allows the model to be trained on different aspects of data, which produces diversity deliberately without resorting to other machine learning algorithms. In general, the more diverse the subsets, the better the predictive performance of LIFE. Hence, the upper bound is necessary to ensure diversity of subset since subset contains almost all observations and its size is very large, making subsets loss diversity. Another parameter cutoff point is set up to balance accuracy and diversity as well.

For binary classification purposes, let  $f_{i}= \sum_{j=1}^M \beta_jf_i^{(j)}$ be individual predicted probability of an ensemble model for $i^{th}$ observation, which is weighted average of predicted probability or log-odds of base learner $f_i^{(j)}$,  where $\sum_{j=1}^M\beta_j=1$ and $\beta_j\geq0$.   The cross-entropy loss is widely used for classification and it can be written as follows for single observation:
\begin{equation} \label{eq:loss_cal}
\begin{gathered}
l(y_i,f_i)=-y_ilog(f_i)-(1-y_i)log(1-f_i). 
\end{gathered}
\end{equation}
By plugging the loss function \eqref{eq:loss_cal} into equation \eqref{eq:dec}, we can write average cross entropy loss of an ensemble method on training set $\{x_i, y_i\}_{i=1,\cdots,N}$  in the probability space as follows: 
\begin{equation} \label{eq:cal_dec}
\begin{gathered}
\sum_{i=1}^N [-y_ilog(f_i)-(1-y_i)log(1-f_i)] \\
=\underbrace{\sum_{i=1}^N\sum_{j=1}^M \beta_j [-y_ilog(f_i^{(j)})-(1-y_i)log(1-f_i^{(j)})]}_\text{accuracy} \\
-\underbrace{\frac{1}{2} \sum_{i=1}^N\sum_{j=1}^M \beta_j \{\frac{y_i-2f_i^{(j)\star}y_i+(f_i^{(j)\star})^2}{[f_i^{(j)\star}(1-f_i^{(j)\star})]^2}\}(f_i- f_{i}^{(j)})^2}_\text{diversity},
\end{gathered}
\end{equation}
\begin{figure}[h]
     \centering
   \subfloat[][Accuracy and Diversity (MIM)]{\includegraphics[width=3.2in]{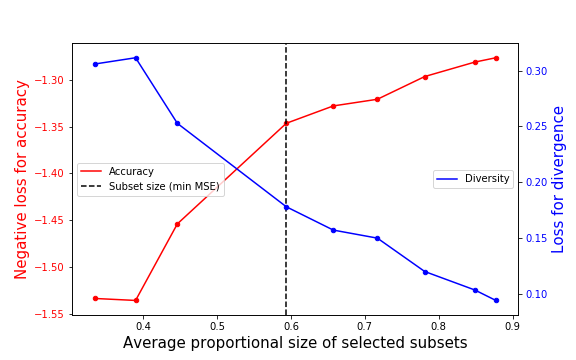}\label{fig:acc_divM}} 
     \subfloat[][Accuracy and Diversity (California Housing)]{\includegraphics[width=3.2in]{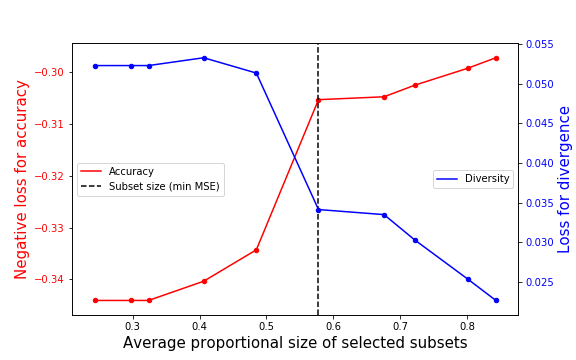}\label{fig:acc_divC}}\\
     \subfloat[][Accuarcy and MSE (MIM)]{\includegraphics[width=3.2in]{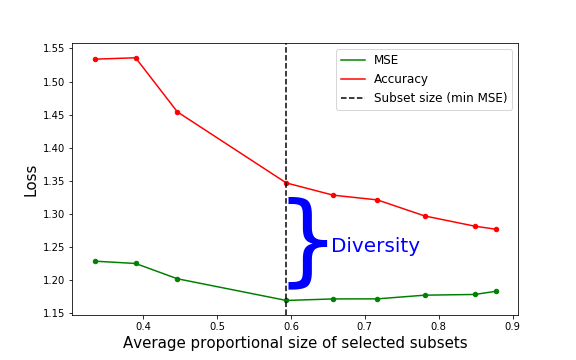}\label{fig:acc_mseM}} 
     \subfloat[][Accuarcy and MSE (California Housing)]{\includegraphics[width=3.2in]{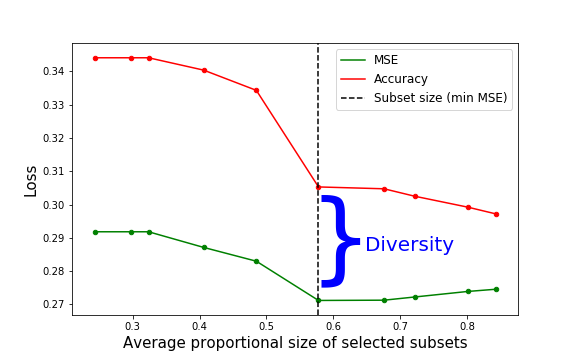}\label{fig:acc_mseC}}
     \caption{Loss decomposition for regression}
     \label{fig:loss_dec}
\end{figure}
\begin{figure}[h]
     \centering
   \subfloat[][Accuracy and Diversity (MIM)]{\includegraphics[width=3.2in]{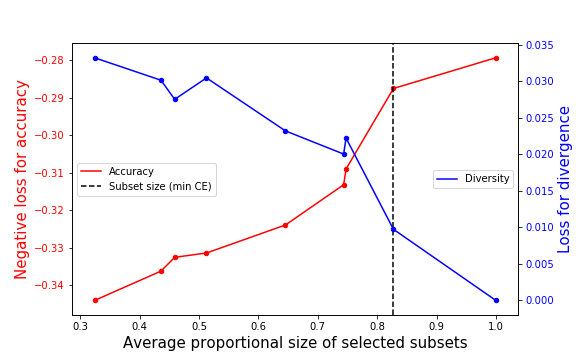}\label{fig:acc_divM_cla}}
     \subfloat[][Accuracy and Diversity (Gamma Telescope)]{\includegraphics[width=3.2in]{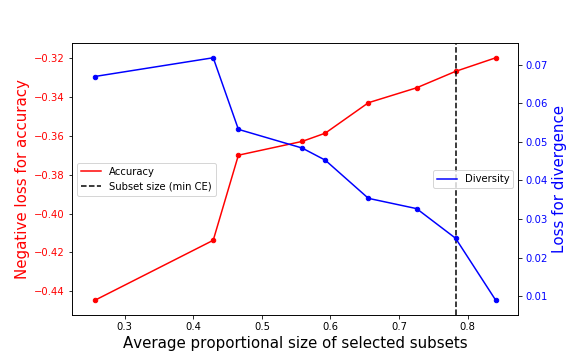}\label{fig:acc_div_cla}} \\
     \subfloat[][Accuarcy and Cross Entropy (MIM)]{\includegraphics[width=3.2in]{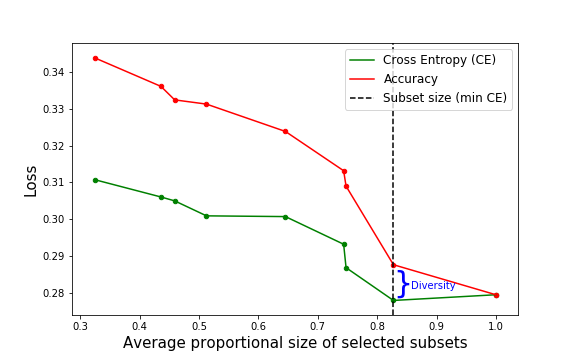}\label{fig:acc_mse_cla}}
     \subfloat[][Accuarcy and Cross Entropy (Gamma Telescope)]{\includegraphics[width=3.2in]{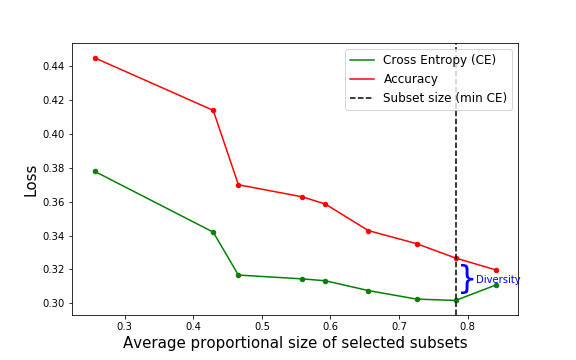}\label{fig:acc_mse_cla}}
     \caption{Loss decomposition for classification}
     \label{fig:loss_dec_cla}
\end{figure}
where $f_i^{(j)\star}$ takes value between $f_i^{(j)}$ and $f_i$.
The term $(f_i-f_i^{(j)})^2$ is a measure of the differences in value between base learner and the ensemble. The cross-entropy loss and its decomposition in the log-odds space is provided in the Appendix \ref{sect.log}.
Unlike diversity term in the regression case, the second term (diversity) in the right-hand side of equation \eqref{eq:cal_dec} also includes the true class label $y_i$ and parameter with unknown value $f_i^{(j)\star}$. However,  the interpretation of decomposition is also clear.  
It shows that a lower average accuracy of individual base learner can be compensated by a higher disagreement with the ensemble, scaled by $\frac{1}{(f_i^{(j)\star})^2}$ if $y_i=1$ or $\frac{1}{(1-f_i^{(j)\star})^2}$ if $y_i=0$ in the probability space. Since $\frac{\beta_j}{(f_i^{(j)\star})^2}$ or $\frac{\beta_j}{(1-f_i^{(j)\star})^2}$ is positive, the more deviance of predicted probability between base learner and an ensemble model implies more diversity. 


We implement the loss decomposition on simulated data (MIM) and real data (California Housing for regression and Gamma Telescope for classification) to LIFE without neural nodes flattening.  It ensembles the predictions from single-hidden-layer NN base learner directly which is the two-stage stacking model averaging method discussed above. As subset size in the sampling step of LIFE impact the strength of diversity, we explore the overall loss, the weighted sum individual base learner accuracy and the ambiguity measure against the average subset size over all the single-hidden-layer NNs in the first step of LIFE.   Here we vary the subset size by controlling the cutoff point $cp$ for linear project. Figures \ref{fig:loss_dec}  and \ref{fig:loss_dec_cla} show the relationship between average subset size (in terms of the proportion of original train data size) and loss for accuracy, loss for diversity and MSE loss in both regression and classification cases.  In plot (a) and (b), the blue curves indicating the ambiguity diversity measure have a decreasing trend when the subset size increases.  This is consistent with the intuition that the larger overlapping the subsets are, the less diverse the base learners are.  On the other hand, the accuracy of the individual base learners is higher when subset size is larger, as the sample is more representative of the whole training set.  Similarly, training on smaller subsets has stronger diversity but leads to lower accuracy.  Plot (c) and (d) illustrate the tradeoff between average accuracy and diversity to minimize total loss, where the dash line shows the optimal subset size achieving the minimum loss.  Combining the results from Section \ref{Connection} and Section \ref{LossDecomp}, we conclude that competitive predictive performance of LIFE benefits from diversity due to data sampling in the first step and the feature ``flattening" and joint estimation in the third step.

\section{Empirical Experiment} \label{sec:ee}
In this section, we conduct multiple empirical experiments via both simulated and real data for regression and classification cases to confirm the competitive performance of LIFE. We have generated multiple datasets with normal distribution and heavy-tailed predictor distribution (Laplace distribution), as well as different function forms to analyze predictive performance and computational efficiency. Other benchmark models including single-hidden-layer NN trained by different optimizer (Local linear approximation and Adam algorithm), and other machine learning algorithms including multi-layer FFNN, Xgboost, and Random Forest are tested on the same data for comparison after extensive hyper-parameter tuning. Local Linear Approximation (LLA) algorithm is a recently proposed method to estimate the weights and biases of single-hidden-layer NN by iterative linear regression and linear approximation of the ReLU activation function \cite{Zeng2020LLA}. The LLA algorithm is distinguished from existing gradient descent algorithms in that it utilizes the Hessian matrix in the same spirit of Fisher scoring algorithm for nonlinear regression models with normal error. The outline of the LLA algorithm is included in the Appendix \ref{sect.lla}.

\subsection{Simulated Data}
\subsubsection{Regression}
For the regression scenario, there are three different function forms including Generalized Additive Model (GAM), Additive Index Model (AIM) and Multiple Index Model (MIM), which are expressed, as follows:

\begin{equation} \label{eq:sim_gam}
\begin{gathered}
GAM: y_i =\beta_1 x_{1i} +\beta_2\sqrt{|x_{2i}|} +\beta_3|x_{3i}|+ \beta_4exp(x_{4i})+\beta_5log(|x_{5i}|)+\beta_6max(1,x_{6i})+\epsilon_i,\\
\beta = \{\beta_1, \cdots, \beta_6\}=\{1.5,\sqrt{5}, 2,4e^{(\frac{-1.5}{7})}, 4log(1.5),-4\},   \epsilon_i  \sim N(0,1), i=1,\dots,N,
\end{gathered}
\end{equation}

\begin{equation} \label{eq:sim_aim}
\begin{gathered}
AIM:  y_i = 2log(|\beta_1x_{1i}+\cdots+\beta_4x_{4i}|)+exp(\frac{\beta_3x_{3i}+\cdots+\beta_6x_{6i}}{9})\\
+max(0,\beta_5x_{5i}+\beta_6x_{6i})+\epsilon_i, \\ 
\beta = \{\beta_1, \cdots, \beta_6\}=\{3,-2.5,2,-1.5,1.5,-1\},
\end{gathered}
\end{equation}

\begin{equation} \label{eq:sim_mim}
\begin{gathered}
MIM:  y_i = exp(\beta_1 x_{1i}+\beta_2 x_{2i})\beta_3x_{3i}+\frac{\beta_4x_{4i}}{1+\beta_5|x_{5i}|}+max(2,\beta_6x_{6i})+\epsilon_i, \\
\beta = \{\beta_1, \cdots, \beta_6\}=\{0.03,-0.025,1,-3,1.5,-2\},
\end{gathered}
\end{equation}
where $N=20k$ and all predictors $\{x_{ji}\}_ {j=1,\cdots,6;i=1,\cdots,N}$ are drawn from Normal or Laplace distribution. 
For regression, the experimental results show mean and standard deviation of $RMSE$, $R^2$ and training time $T$ over five replications in Tables \ref{tab:sim_reg1} and  \ref{tab:sim_reg2}, while log-loss and AUC are used as a performance metric in the classification case as shown in Tables \ref{tab:sim_cal1} and  \ref{tab:sim_cal2}. Bayesian Optimization allows us to jointly tune more parameters with fewer experiments and find better values, so we implement it to perform extensive hyper–parameter tuning on all the algorithm.  The important hyper-parameters for LIFE include the number of iterations, the number of neurons in each iteration, upper and lower bound. We marked optimal results that have won the campaign in red.

\begin{table}[h]
\centering 
\scalebox{0.85}{
  \begin{threeparttable}
\caption{Regression on Simulated Data (Normal Distribution)} 
\begin{tabular}{l c c c c c c | c c c} 
\hline\hline 
 Model & Metric & Oracle & LIFE(LLA) & LLA & LIFE(Adam) & Adam &FFNN  &Xgboost & RF
\\ [0.5ex]
\hline 
 & &  &\textcolor{red}{\textbf{1.046}} & 1.062 & 1.063 & 1.149 &1.104 & 1.065 & 1.906 \\  [-1.5ex]
 &\raisebox{1.5ex}{$RMSE$} & \raisebox{1.5ex}{$1.000$} &(0.012) & (0.040) & (0.016) & (0.023)&(0.025)  & (0.020)& (0.037) \\  
GAM  & &  &\textcolor{red}{\textbf{0.962}} & 0.961 & 0.961 & 0.954& 0.957 & 0.960 & 0.873\\[-1.5ex]
 &\raisebox{1.5ex}{$R^2$} & \raisebox{1.5ex}{$0.965$} & (0.001) & (0.003) & (0.001) & (0.002)& (0.002) & (0.002)& (0.005) \\ 
 &$T$ & N/A & 75s & 120s & 46s &  55s & 81s  & \textcolor{red}{\textbf{8s}}& 44s \\
\hline 
 & &  &\textcolor{red}{\textbf{1.077}} & 1.092 & 1.153 & 1.202 & 1.159 & 2.606 & 3.285 \\  [-1.5ex]
 &\raisebox{1.5ex}{$RMSE$} & \raisebox{1.5ex}{$1.000$} &(0.018) & (0.032) & (0.029) & (0.040)&(0.033)  & (0.173)& (0.161) \\  
AIM  & &  &\textcolor{red}{\textbf{0.981}} & 0.980 & 0.978 & 0.975&0.978  & 0.886 & 0.828\\[-1.5ex]
 &\raisebox{1.5ex}{$R^2$} & \raisebox{1.5ex}{$0.984$} & (0.001) & (0.001) & (0.001) & (0.004)&(0.001)  & (0.010)& (0.010) \\ 
 &$T$ & N/A & \textcolor{red}{\textbf{7s}} & 51s & 17s & 40s & 46s  & 27s & 48s \\
\hline
 & &  &\textcolor{red}{\textbf{1.028}} & 1.043 & 1.036 & 1.047 &1.037  & 1.057& 1.139 \\  [-1.5ex]
 &\raisebox{1.5ex}{$RMSE$} & \raisebox{1.5ex}{$1.000$} &(0.005) & (0.003) & (0.006) & (0.004) & (0.009) &(0.004) & (0.011) \\  
MIM  & &  &\textcolor{red}{\textbf{0.933}} & 0.931 & 0.931 & 0.930 & 0.931 & 0.929 & 0.918\\[-1.5ex]
 &\raisebox{1.5ex}{$R^2$} & \raisebox{1.5ex}{$0.937$} & (0.003) & (0.003) & (0.003) & (0.003) & (0.007) & (0.003) & (0.008)\\ 
 &$T$ & N/A & 29s & 313s & 81s & 45s & 51s  &\textcolor{red}{\textbf{14s}}& 44s \\
\hline\hline 
\end{tabular}
 \begin{tablenotes}
      \small
      \item Note: On the left side of vertical line, all columns represent just single-hidden-layer NNs optimized by LIFE (LLA), LLA, LIFE (Adam) and Adam, while there are three state-of-art machine learning methods including FFNN, Xgboost, and Random Forest (RF) on the right side. FFNN is multi-hidden-layer feed forward NN, where range for number of hidden layers is from two to four. In addition, Xgboost and RF are two tree-based ensemble methods. The figures inside parenthesis indicate the standard deviation of metrics, and time represents training time of one replication. The numbers colored in red represents the optimal results of this metric.
    \end{tablenotes}
\label{tab:sim_reg1}
 \end{threeparttable}
}
\end{table}

\begin{table}[h]
\centering 
\scalebox{0.85}{
  \begin{threeparttable}
\caption{Regression on Simulated Data (Laplace Distribution)} 
\begin{tabular}{l c c c c c c | c c c} 
\hline\hline 
 Model & Metric & Oracle & LIFE(LLA) & LLA & LIFE(Adam) & Adam &  FFNN & Xgboost & RF
\\ [0.5ex]
\hline 
 & &  &\textcolor{red}{\textbf{1.166}} & 1.258 & 1.427 & 1.629 &1.460 & 1.439  & 3.242\\  [-1.5ex]
 &\raisebox{1.5ex}{$RMSE$} & \raisebox{1.5ex}{$1.000$} &(0.079) & (0.091) & (0.192) & (0.100) & (0.205)  & (0.256) &(0.226) \\  
GAM  & &  &\textcolor{red}{\textbf{0.989}} & 0.987 & 0.984 & 0.979 & 0.983 &0.983 &0.920 \\[-1.5ex]
 &\raisebox{1.5ex}{$R^2$} & \raisebox{1.5ex}{$0.992$} & (0.001) & (0.001) & (0.003) & (0.002) & (0.004) &(0.005)  & (0.005) \\ 
 &$T$ & N/A & 48s & 84s & 19s & 96s & 75s  & \textcolor{red}{\textbf{5s}}   & 20s \\
\hline 
 & &  &\textcolor{red}{\textbf{1.057}} & 1.084 & 1.122 & 1.192 & 1.166  & 2.031  & 2.491\\  [-1.5ex]
 &\raisebox{1.5ex}{$RMSE$} & \raisebox{1.5ex}{$1.000$} &(0.026) & (0.040) & (0.055) & (0.084)& (0.077)  & (0.099) & (0.204) \\  
AIM  & &  &\textcolor{red}{\textbf{0.966}} & 0.964 & 0.962 & 0.957 & 0.959 &0.876 & 0.804\\[-1.5ex]
 &\raisebox{1.5ex}{$R^2$} & \raisebox{1.5ex}{$0.968$} & (0.004) & (0.005) & (0.003) & (0.004) & (0.004)  & (0.012)  & (0.035)\\ 
 &$T$ & N/A & 10s & 61s & 22s & 59s & 26s & \textcolor{red}{\textbf{8s}}  & 56s\\
\hline
 & &  &\textcolor{red}{\textbf{1.060}} & 1.110 & 1.060 & 1.096 & 1.088 &1.099 & 1.555 \\  [-1.5ex]
 &\raisebox{1.5ex}{$RMSE$} & \raisebox{1.5ex}{$1.000$} &(0.017) & (0.065) & (0.018) & (0.026) & (0.018) & (0.013) & (0.081)\\  
MIM  & &  &\textcolor{red}{\textbf{0.965}} & 0.961 & 0.965 & 0.962 & 0.963 & 0.962  & 0.925\\[-1.5ex]
 &\raisebox{1.5ex}{$R^2$} & \raisebox{1.5ex}{$0.969$} & (0.001) & (0.005) & (0.001) & (0.002) &(0.001) & (0.001)  & (0.008)\\ 
 &$T$ & N/A & 20s & 195s & 39s & 42s & 32s  &\textcolor{red}{\textbf{6s}}  & 43s\\
\hline\hline 
\end{tabular}
 \begin{tablenotes}
      \small
      \item Note: On the left side of vertical line, all columns represent just single-hidden-layer NNs optimized by LIFE (LLA), LLA, LIFE (Adam) and Adam, while there are three state-of-art machine learning methods including FFNN, Xgboost, and Random Forest (RF) on the right side. FFNN is multi-hidden-layer feed forward NN, where range for number of hidden layers is from two to four. In addition, Xgboost and RF are two tree-based ensemble methods. The figures inside parenthesis indicate the standard deviation of metrics, and time represents training time of one replication. The numbers colored in red represents the optimal results of this metric.
    \end{tablenotes}
\label{tab:sim_reg2}
 \end{threeparttable}
}
\end{table}

As illustrated in Tables  \ref{tab:sim_reg1} and  \ref{tab:sim_reg2}, the result is predictable regardless of the distribution predictors drawn. LIFE algorithm with LLA optimizer achieves higher accuracy among all methods in terms of predictive performance on the test set. The values of two metrics from LIFE (LLA) are close to oracle values, which implies LIFE performs well in the data with a smoothing response surface. If we compare results from one hidden-layer FFNNs trained by LIFE algorithm with either LLA or Adam base learners and non-ensemble algorithms of LLA or Adam, LIFE always outperforms the relevant optimization methods used to train single hidden-layer FFNN as a whole due to the generated diversity of data sampling. In addition, the performance of LIFE also depends on the strength of individual NN base learner, which can be easily spotted in the Tables \ref{tab:sim_reg1} and \ref{tab:sim_reg2} that LIFE (LLA) outperforms LIFE(Adam). From the perspective of computational efficiency, LIFE algorithm also shows some advantages over other single-hidden-layer NN training algorithms. In general, LIFE algorithm can not only boost predictive performance of one-hidden layer NN, but also speed up training, especially with respect to wide NN with large hidden layer dimension.

\subsubsection{Classification}
For classification case, the functional forms in simulation setup are similar to the ones in regression case except that the coefficients are a little bit different. Detailed information on formulas can be found in Appendix \ref{sect.sim_model}. Similar to the setup in regression case, we choose $N=20k$ and all predictors are drawn from either Normal or Laplace distribution. The response variable is sampled from Bernoulli distribution with probability calculated using the logit link function.

\begin{table}[h] 
\centering 
\scalebox{0.85}{
  \begin{threeparttable}
\caption{Classification on Simulated Data (Normal Distribution)} 
\begin{tabular}{l c c c c c c | c c c} 
\hline\hline 
 Model & Metric & Oracle &  LIFE(LLA) & LLA & LIFE(Adam) & Adam &  FFNN & Xgboost & RF \\ [0.5ex]
%

\hline 

                                        &                                           & $  0.984  $   &   0.974   &  0.968   &  0.973   &  0.945  & 0.973  & \textcolor{red}{\textbf{0.975}}   &  0.964  \\  [-1.5ex]
                                       &\raisebox{1.5ex}{$AUC$}      & $(0.002)$ & (0.002) & (0.003) & (0.002) & (0.003) &  (0.002)  &  (0.002) &  (0.002)\\  
\raisebox{1.5ex}{GAM}    &                                           & $  0.113  $   &  0.146 &  0.158   &  0.148   &  0.247   &  0.152  &  \textcolor{red}{\textbf{0.141}}  & 0.166     \\[-1.5ex]
                                       &\raisebox{1.5ex}{$logloss$} & $ (0.005) $& (0.004) & (0.010) & (0.005) & (0.029) &  (0.005)  & (0.006) & 0.006     \\ 
[0.5ex] 

\hline

                                       &                                           & $  0.836   $     & 0.751   &  0.535   & \textcolor{red}{\textbf{0.752}}    & 0.742   &  \textcolor{red}{\textbf{0.752}}   &  0.687  &  0.715   \\  [-1.5ex]
                                       &\raisebox{1.5ex}{$AUC$}     & $(0.008)$ & (0.010) & (0.037) & (0.010) & (0.010) & (0.010)   & (0.010) &  (0.011)  \\  
\raisebox{1.5ex}{AIM}    &                                           & $  0.359   $     &   \textcolor{red}{\textbf{0.422}}   &  0.485    &  0.428   & 0.448   & 0.431   & 0.457  &  0.447  \\[-1.5ex]
                                       &\raisebox{1.5ex}{$logloss$} & $ (0.005) $  & (0.003) & (0.005) & (0.006) & (0.010) & (0.004)  &  (0.001)  &  (0.002)     \\ 
[0.5ex] 

\hline

                                       &                                           & $ 0.888    $      & \textcolor{red}{\textbf{0.839}}    & 0.838    &  0.838   & 0.830   &  \textcolor{red}{\textbf{0.839}}  & 0.837   &  0.834   \\  [-1.5ex]
                                       &\raisebox{1.5ex}{$AUC$}     & $(0.005)$  & (0.008) & (0.007) & (0.008) & (0.008) &  (0.008)  & (0.008) & (0.007)  \\  
\raisebox{1.5ex}{MIM}    &                                           & $ 0.331    $   & \textcolor{red}{\textbf{0.386}}    & 0.388    & 0.387    & 0.399  &  0.387    & 0.388  &  0.391  \\[-1.5ex]
                                       &\raisebox{1.5ex}{$logloss$} & $ (0.010) $  & (0.010) & (0.009) & (0.010) & (0.011) & (0.009)  &  (0.010) &  (0.009)  \\

\hline\hline 
\end{tabular}
 \begin{tablenotes}
      \small
      \item Note: On the left side of vertical line, all columns represent just single-hidden-layer NNs optimized by LIFE (LLA), LLA, LIFE (Adam) and Adam, while there are three state-of-art machine learning methods including FFNN, Xgboost, and Random Forest (RF) on the right side. FFNN is multi-hidden-layer feed forward NN, where range for number of hidden layers is from two to four. In addition, Xgboost and RF are two tree-based ensemble methods. The figures inside parenthesis indicate the standard deviation of metrics, and time represents training time of one replication. The numbers colored in red represents the optimal results of this metric.
    \end{tablenotes}
\label{tab:sim_cal1}
 \end{threeparttable}
}
\end{table}

\begin{table}[h]
\centering 
\scalebox{0.85}{
  \begin{threeparttable}
\caption{Classification on Simulated Data (Laplace Distribution)} 
\begin{tabular}{l c c c c c c | c c c} 
\hline\hline 
 Model & Metric & Oracle  &  LIFE(LLA) & LLA & LIFE(Adam) & Adam & FFNN & Xgboost & RF \\ [0.5ex]

\hline 

                                        &                                           & $  0.996   $  &  0.990   &  0.984   &  0.990   & 0.857   &  0.976   &  \textcolor{red}{\textbf{0.991}}   &  0.979   \\  [-1.5ex]
                                       &\raisebox{1.5ex}{$AUC$}     & $(0.000)$ & (0.001) & (0.003) & (0.001) & (0.012) &  (0.002)  &  (0.001)   & (0.002)   \\  
\raisebox{1.5ex}{GAM}  &                                           & $  0.051   $    &  0.081   & 0.100    &  0.081    &  0.252   &  0.133  &  \textcolor{red}{\textbf{0.079}}    &  0.120   \\[-1.5ex]
                                       &\raisebox{1.5ex}{$logloss$} & $ (0.002) $  & (0.003) & (0.006) & (0.003) & (0.012) &  (0.004)   &  (0.004)  &  (0.003) \\
 [0.5ex] 

\hline

                                       &                                           & $  0.866   $    &  \textcolor{red}{\textbf{0.802}}   &  0.797   &  \textcolor{red}{\textbf{0.802}}   &  0.795  &  \textcolor{red}{\textbf{0.802}}   &  0.737   &  0.743   \\  [-1.5ex]
                                       &\raisebox{1.5ex}{$AUC$}     & $(0.012)$ & (0.018) & (0.020) & (0.018) & (0.021) & (0.017)  &  (0.014) & (0.015)    \\  
\raisebox{1.5ex}{AIM}  &                                           & $   0.230  $ &  \textcolor{red}{\textbf{0.270}}   & 0.274    & 0.271    &  0.281  &  0.273   & 0.298 &  0.296    \\[-1.5ex]
                                       &\raisebox{1.5ex}{$logloss$} & $ (0.006) $  & (0.008) & (0.012) & (0.008) & (0.007) &  (0.007)  &  (0.004) &  (0.004)   \\ 
 [0.5ex] 

\hline

                                       &                                           & $   0.958  $    &  0.939   &  0.938   & \textcolor{red}{\textbf{0.940}}    & 0.920   &  0.937   &  0.938  &   0.934   \\  [-1.5ex]
                                       &\raisebox{1.5ex}{$AUC$}     & $(0.003)$  & (0.004) & (0.003) & (0.004) & (0.003) & (0.002)   &  (0.004) &  (0.004)    \\  
\raisebox{1.5ex}{MIM}  &                                           & $  0.225   $    &  \textcolor{red}{\textbf{0.268}}   &  0.272   & \textcolor{red}{\textbf{0.268}}    &  0.421   &   0.276   &  0.271 &   0.279   \\[-1.5ex]
                                       &\raisebox{1.5ex}{$logloss$} & $ (0.011) $  & (0.012) & (0.011) & (0.012) & (0.075) & (0.010)   &  (0.011) &  (0.010)   \\

\hline\hline 
\end{tabular}
\begin{tablenotes}
      \small
      \item Note: On the left side of vertical line, all columns represent just single-hidden-layer NNs optimized by LIFE (LLA), LLA, LIFE (Adam) and Adam, while there are three state-of-art machine learning methods including FFNN, Xgboost, and Random Forest (RF) on the right side. FFNN is multi-hidden-layer feed forward NN, where range for number of hidden layers is from two to four. In addition, Xgboost and RF are two tree-based ensemble methods. The figures inside parenthesis indicate the standard deviation of metrics, and time represents training time of one replication. The numbers colored in red represents the optimal results of this metric.
    \end{tablenotes}
\label{tab:sim_cal2}
 \end{threeparttable}
}
\end{table}


Table \ref{tab:sim_cal1} and \ref{tab:sim_cal2} show the simulation results from binary scenario, where data are generated from Normal distribution and Laplace distribution, respectively. Similar to the results in regression case, LIFE(LLA) has won the campaign in four out of six functional forms. LIFE(Adam) also performs pretty well especially in Table \ref{tab:sim_cal2} for Laplace Distribution. For data drawn from Normal Distribution shown in Table \ref{tab:sim_cal1}, LIFE(Adam) is also quite close to the optimal result. 
Furthermore, there is a strong evidence to show LIFE algorithm does improve the performance of base learners with larger AUC, smaller log-loss and smaller standard errors of both metrics. Even if the base learner is not strong enough, like Adam for GAM and MIM in Laplace Distribution (Table \ref{tab:sim_cal2}), with which AUC or log-loss or both has large standard error, the ensemble approach in LIFE(Adam) can dramatically reduce the variance. XGboost ranks at top for GAM in both distributions, however, the differences between LIFE and XGboost are negligible with only 0.1\% of difference in AUC and 3\% of difference in log-loss.

\subsection{Real Data}
Besides implementing LIFE algorithm on simulated data, we also tested it on 7 public datasets for regression and eight datasets for classification and compared it with other benchmark models including single-hidden-layer FFNN and Xgboost. All data sets are split into $80\%$  training data and $20\%$ testing data with 10 different random seeds, which yield results over $10$ replications. For all the datasets, we transformed categorical variables into dummy variables and standardized the continuous variables, so that the mean and the variance of each continuous variable are equal to $0$ and $1$, respectively. A detailed description of all datasets and corresponding data preprocessing steps are outlined in the Appendix \ref{sect.data}.

\subsubsection{Regression} 
The experiment results averaged over ten replications are reported in Table \ref{tab:real_reg}, including root mean squared error ($RMSE$), R-squared ($R^2$) and training time ($T$).
\begin{table}
\centering 
\scalebox{0.85}{
 \begin{threeparttable}
\caption{Regression on Real Data} 
\begin{tabular}{l c c c c c | c c c} 
\hline\hline 
 Model & Metric & LIFE(LLA) & LLA & LIFE(Adam) & Adam & FFNN & Xgboost &RF
\\ [0.5ex]
\hline 
 &  &\textcolor{red}{\textbf{2.112}} & 2.162 &  2.137 & 2.179 & 2.195 & 2.169 & 2.170 \\  [-1.5ex]
 &\raisebox{1.5ex}{$RMSE$}  & (0.077) & (0.074) & (0.071) & (0.108)& (0.133) & (0.068)  & (0.071)\\  
Abalone  &   &\textcolor{red}{\textbf{0.570}} & 0.549 &  0.560 & 0.543 & 0.534 & 0.546 & 0.546\\[-1.5ex]
 &\raisebox{1.5ex}{$R^2$}  & (0.031) & (0.031) & (0.029) & (0.047) & (0.057)& (0.028) & (0.029) \\ 
 &$T$  & 2s & 289s &\textcolor{red}{\textbf{1s}} & 35s & 17s  & \textcolor{red}{\textbf{1s}} &  \textcolor{red}{\textbf{1s}} \\
\hline 

 &  &\textcolor{red}{\textbf{0.225}} & 0.288 & 0.276 & 0.293 & 0.270 &0.238 & 0.343\\  [-1.5ex]
 &\raisebox{1.5ex}{$RMSE$}  & (0.020) & (0.021) & (0.026) & (0.018) & (0.034) &(0.016)& (0.023)\\  
Airfoil  &   &\textcolor{red}{\textbf{0.949}} & 0.917 & 0.923 & 0.913 & 0.925 &0.943 & 0.881\\[-1.5ex]
 &\raisebox{1.5ex}{$R^2$}  &  (0.009) & (0.012) & (0.014) & (0.011) &(0.019)  & (0.008) & (0.016)\\ 
 &$T$  &  \textcolor{red}{\textbf{1s}} & 14s &  \textcolor{red}{\textbf{1s}} & 8s &  \textcolor{red}{\textbf{1s}} & \textcolor{red}{\textbf{1s}} & \textcolor{red}{\textbf{1s}} \\

\hline 
 &  &\textcolor{red}{\textbf{1.193}} & 1.220 & 1.210 & 1.223 &1.255 &1.214 & 1.204\\  [-1.5ex]
 &\raisebox{1.5ex}{$RMSE$}  & (0.066) & (0.045) &(0.068) & (0.096) & (0.125) & (0.101) & (0.082)\\  
Aquatic Toxicity &   &\textcolor{red}{\textbf{0.484}} & 0.461 & 0.469 & 0.455 & 0.425 &0.464 & 0.473\\[-1.5ex]
 &\raisebox{1.5ex}{$R^2$}  & (0.058) & (0.039) & (0.060) & (0.088) & (0.117)  &(0.089) & (0.071)\\ 
 &$T$  &  \textcolor{red}{\textbf{1s}} & 41s &  \textcolor{red}{\textbf{1s}} & 2s & \textcolor{red}{\textbf{1s}}  &\textcolor{red}{\textbf{1s}} & \textcolor{red}{\textbf{1s}}\\

\hline 
 &  &41.98 & 47.66 & 45.34 & 50.74 & 45.99 &\textcolor{red}{\textbf{41.71}} & 66.38\\  [-1.5ex]
 &\raisebox{1.5ex}{$RMSE$}  &(1.299) & (0.946) & (1.103) & (1.601) &(1.737)  & (1.184)  &(1.257)\\  
Bike Sharing &   &0.946 & 0.930 & 0.937 & 0.922 & 0.935 &\textcolor{red}{\textbf{0.947}} & 0.866\\[-1.5ex]
 &\raisebox{1.5ex}{$R^2$}  &(0.003) & (0.003) & (0.003) & (0.005) & (0.005) & (0.003) & (0.005)\\ 
 &$T$  & 85s & 2194s & 81s & 266s & 114  &13s &\textcolor{red}{\textbf{12s}} \\

\hline 
 &  &0.500 & 0.527 & 0.521 & 0.535 & 0.519 &\textcolor{red}{\textbf{0.479}} & 0.486\\  [-1.5ex]
 &\raisebox{1.5ex}{$RMSE$}  &(0.012) & (0.011) & (0.009) & (0.008) & (0.011) & (0.007) & (0.007)\\  
California Housing  &   &0.812 & 0.791 & 0.796 & 0.784 &0.797  &\textcolor{red}{\textbf{0.825}} & 0.822\\[-1.5ex]
 &\raisebox{1.5ex}{$R^2$}  &(0.009) & (0.008) & (0.007) & (0.007) & (0.008) & (0.005) &(0.005)\\ 
 &$T$  & 13s & 108s & \textcolor{red}{\textbf{8s}} & 74s & 40s &12s & 40s\\

\hline 
 &  &3.865 & 4.111 & 4.081 & 4.133 &  4.067 &\textcolor{red}{\textbf{3.786}} & 3.951 \\  [-1.5ex]
 &\raisebox{1.5ex}{$RMSE$}  &(0.034) & (0.049) & (0.047) & (0.071) & (0.037) &  (0.022) & (0.016)\\  
CASP  &   &0.601 & 0.548 & 0.555 & 0.543 & 0.558 &\textcolor{red}{\textbf{0.617}}  & 0.583 \\[-1.5ex]
 &\raisebox{1.5ex}{$R^2$}  & (0.007) & (0.011) & (0.010) & (0.016) & (0.008) & (0.004) & (0.004)\\ 
 &$T$  & 216s & 2881s & \textcolor{red}{\textbf{54s}} & 339s & 83s &60s & 56s\\

\hline 
 &  &\textcolor{red}{\textbf{0.007}} & 0.011 &  0.008 & 0.009 &0.009  &0.011 &0.013 \\  [-1.5ex]
 &\raisebox{1.5ex}{$RMSE$}  & (0.000) & (0.001) & (0.000) & (0.001) & (0.001)  &(0.000) & (0.000)\\  
Electrical Grid &   &\textcolor{red}{\textbf{0.959}} & 0.908 &  0.949 & 0.944 & 0.945  &0.908 & 0.877\\[-1.5ex]
 &\raisebox{1.5ex}{$R^2$}  & (0.003) & (0.016) &  (0.003) & (0.009) & (0.011)  &(0.003) & (0.004)\\ 
 &$T$  & 25s & 27s & \textcolor{red}{\textbf{4s}} & 27s &17s &16s & 18s\\

\hline\hline 
\end{tabular}
 \begin{tablenotes}
      \small
      \item Note: On the left side of vertical line, all columns represent just single-hidden-layer NNs optimized by LIFE (LLA), LLA, LIFE (Adam) and Adam, while there are three state-of-art machine learning methods including FFNN, Xgboost, and Random Forest (RF) on the right side. FFNN is multi-hidden-layer feed forward NN, where range for number of hidden layers is from two to four. In addition, Xgboost and RF are two tree-based ensemble methods. Thefigures inside parenthesis indicate the standard deviation of metrics, and time represents training time of one replication. The numbers colored in red represents the optimal results of this metric.
    \end{tablenotes}
\label{tab:real_reg}
 \end{threeparttable}
}
\end{table}
As observed in Table \ref{tab:real_reg}, LIFE (LLA) is ranked as the best algorithm in the four datasets. For the remaining three datasets, LIFE(LLA) is still the second or third best algorithm among all models with a close or slightly worse predictive performance than optimal one (Xgboost or Random Forest), which implies that the LIFE algorithm is competitive with other state-of-art machine learning algorithms. In addition, there is an average $4.6\%$ or $1.8\%$  improvement in R-square of all real datasets when single-hidden-layer NN is trained by LIFE (LLA or Adam) instead of other optimization methods (LLA or Adam), which is consistent with the conclusion made from experiment in the simulated data. It is also worth mentioning that computation efficiency of NN training has been significantly boosted for almost all dataset via LIFE compared with traditional NN training methods. In particular, if we take a look at the largest real dataset CASP, training time of NN via LIFE reduces to $216$ seconds from $2881$ seconds or to $54$ seconds from $339$ seconds when we use LLA as optimizer or Adam respectively, which is almost more than six times faster. Although tree-based ensemble methods such as Xgboost and Random Forest show strong predictive power in some datasets, they are still black-box models, and they are hard to interpret. The biggest advantage of our proposed algorithm LIFE is that it preserves the interpretability of model, which is still single-hidden-layer NN with very strong predictive performance and boosted computation efficiency.

\subsubsection{Classification}
\begin{table}
\centering 
\scalebox{0.85}{
  \begin{threeparttable}
\caption{Classification on Real Data} 
\begin{tabular}{l c c c c c | c c c } 
\hline\hline 
 Data & Metric & LIFE(LLA)  & LLA & LIFE(Adam) & Adam & FFNN & Xgboost & RF
\\ [0.5ex]
\hline 

                                                       &                                           & 0.796       &  0.790     &  0.796   &  0.793    &   0.794   &  0.799  &   \textcolor{red}{\textbf{0.800}}   \\  [-1.5ex]
                                                       &\raisebox{1.5ex}{$AUC$}   & (0.006)  & (0.007)  & (0.006) & (0.006) &  (0.006)  &  (0.008)  &   (0.005)    \\  
\raisebox{1.5ex}{Bank Marketing}  &                                           &   0.288       &  0.292     &  0.289    &  0.294    &  0.288    &  0.288  &  \textcolor{red}{\textbf{0.286}}    \\  [-1.5ex]
                                                      &\raisebox{1.5ex}{$logloss$}& (0.003)   & (0.003)  & (0.004) & (0.003) &  (0.003)  &  (0.003) &  (0.003)       \\ 
[0.5ex]  

\hline 

                                          &                                           &   \textcolor{red}{\textbf{0.997}}       &  0.991     & 0.996    &  0.994    &  0.996   &  0.992  &  0.993   \\  [-1.5ex]
                                       &\raisebox{1.5ex}{$AUC$}   & (0.002)  & (0.009)  & (0.004) & (0.005) & (0.004)   &  (0.006) &  (0.006)       \\  
\raisebox{1.5ex}{Breast Cancer}  &                                 &  0.096       &  0.143     &  \textcolor{red}{\textbf{0.084}}   &  0.340    &  0.099  &  0.114  &  0.090  \\  [-1.5ex]
      \raisebox{1.5ex}{Wisconsin}  &\raisebox{1.5ex}{$logloss$}& (0.015)   & (0.088)  & (0.027) & (0.207) &  (0.014)   &  (0.042)   &  (0.019)      \\ 
[0.5ex]  

\hline 

                                                  &                                           &  0.912     &  0.904     &  \textcolor{red}{\textbf{0.914}}    &  0.891    &  0.912   &  0.911  &  0.911    \\  [-1.5ex]
                                                  &\raisebox{1.5ex}{$AUC$}   & (0.001)  & (0.002)  & (0.001) & (0.003) &  (0.001)  &  (0.001)  &  (0.001)      \\  
\raisebox{1.5ex}{Higgs Boson}  &                                           &  0.353    &  0.368     &  \textcolor{red}{\textbf{0.348}}    &  0.399    &  0.353    &   0.355   &   0.357   \\  [-1.5ex]
                                                 &\raisebox{1.5ex}{$logloss$} & (0.001)  & (0.004)  & (0.001) & (0.005) &  (0.001)  &  (0.001)  &  (0.001)     \\ 
[0.5ex]  

\hline 

                                                    &                                           &  0.853        &  0.841     &  \textcolor{red}{\textbf{0.860}}    &  0.848    & 0.858  & 0.858  &   0.854   \\  [-1.5ex]
                                                    &\raisebox{1.5ex}{$AUC$}   & (0.005) & (0.003)  & (0.003) & (0.006) &  (0.003)  & (0.003)  &   (0.003)      \\  
\raisebox{1.5ex}{Home Lending}  &                                           &  \textcolor{red}{\textbf{0.046}}    &  0.047     &  \textcolor{red}{\textbf{0.046}}    &  0.047    &   \textcolor{red}{\textbf{0.046}}   &   \textcolor{red}{\textbf{0.046}}  &   \textcolor{red}{\textbf{0.046}}     \\  [-1.5ex]
                                                    &\raisebox{1.5ex}{$logloss$}  & (0.001)   & (0.001)  & (0.001) & (0.001)  &  (0.001)  & (0.001)   &  (0.001)     \\ 
 [0.5ex]  

\hline 

                                          &                                           &  \textcolor{red}{\textbf{0.943}}      &  0.928     & 0.938     & 0.909     &  0.938   &  0.936  &  0.937    \\  [-1.5ex]
                                 &\raisebox{1.5ex}{$AUC$}   & (0.002)  & (0.003)  & (0.003) & (0.008) &  (0.003)  &  (0.002)  &  (0.003)      \\  
\raisebox{1.5ex}{MAGIC Gamma}  &                                &  \textcolor{red}{\textbf{0.275}}       &  0.311     & 0.288  &  0.371    &  0.287  &  0.291 &   0.296   \\  [-1.5ex]
    \raisebox{1ex}{Telescope} &\raisebox{1.5ex}{$logloss$}& (0.007)   & (0.007)  & (0.007) & (0.019) &  (0.006)   &  (0.005)   &   (0.005)   \\ 
[0.5ex]  

\hline 

                                          &                                           &   \textcolor{red}{\textbf{1.000}}      &   \textcolor{red}{\textbf{1.000}}   &    \textcolor{red}{\textbf{1.000}}   &   \textcolor{red}{\textbf{1.000}}    &   \textcolor{red}{\textbf{1.000}}   &    \textcolor{red}{\textbf{1.000}}  &  \textcolor{red}{\textbf{1.000}} \\  [-1.5ex]
                                          &\raisebox{1.5ex}{$AUC$}   & (0.000)   & (0.000)  & (0.000) & (0.000) &  (0.000)  &  (0.000) &  (0.000)     \\  
\raisebox{1.5ex}{Mushroom}  &                                           &  0.001      &  \textcolor{red}{\textbf{0.000}}     &  0.001    &  0.024    & 0.004    & 0.004 &   0.005    \\  [-1.5ex]
                                          &\raisebox{1.5ex}{$logloss$}    & (0.000)  & (0.000)  & (0.001) & (0.036) &  (0.003)  &  (0.001) &   (0.000)      \\ 
[0.5ex]  
\hline 

                                          &                                           &  \textcolor{red}{\textbf{0.998}}        &  0.997     &  \textcolor{red}{\textbf{0.998}}    &  0.995    &  0.997    & \textcolor{red}{\textbf{0.998}}  &  0.995    \\  [-1.5ex]
                                          &\raisebox{1.5ex}{$AUC$}   & (0.001)   & (0.001)  & (0.001) & (0.002) &  (0.001)  &  (0.000) &    (0.001)      \\  
\raisebox{1.5ex}{Ringnorm}  &                                           &   \textcolor{red}{\textbf{0.049}}       &  0.086     &  0.051   &  0.147    &  0.075    &  0.057  &    0.162   \\  [-1.5ex]
                                          &\raisebox{1.5ex}{$logloss$}    & (0.006)  & (0.020)  & (0.007) & (0.009) & (0.015)  &  (0.004) &  (0.006)      \\ 
[0.5ex]  

\hline\hline 
\end{tabular}
 \begin{tablenotes}
      \small
      \item Note: On the left side of vertical line, all columns represent just single-hidden-layer NNs optimized by LIFE (LLA), LLA, LIFE (Adam) and Adam, while there are three state-of-art machine learning methods including FFNN, Xgboost, and Random Forest (RF) on the right side. FFNN is multi-hidden-layer feed forward NN, where range for number of hidden layers is from two to four. In addition, Xgboost and RF are two tree-based ensemble methods. The figures inside parenthesis indicate the standard deviation of metrics, and time represents training time of one replication. The numbers colored in red represents the optimal results of this metric.
    \end{tablenotes}
\label{tab:real_cla}
 \end{threeparttable}
}
\end{table}
In the classification case, original LLA algorithm is not stable, since it involves matrix inversion. We added a ridge parameter into the matrix inversion and treat it as a hyper-parameter in LLA algorithm. Experiments have shown adding ridge parameter in LLA can give better and more stable prediction than not adding ridge parameter. After testing LLA and LIFE(LLA) with or without ridge parameter, we further choose the best one for the performance.

Table \ref{tab:real_cla} presents similar patterns in real data analyses as in simulation studies with LIFE(LLA) and LIFE(Adam) taking turns to occupy the dominant position for most of the data sets.  
LIFE performs much better than Xgboost and Random Forest (RF) in most experiments. For example, with Breast Cancer Wisconsin data, log-loss in LIFE(Adam) is 26.3\% lower than that in Xgboost, and with MAGIC Gamma Telescope data, log-loss has dropped by 7\% from Random Forest to LIFE(LLA). The performance of some datasets, such as Bank Marketing data, where LIFE cannot outperform Xgboost or RF, however, the performance is competitive, with only 0.5\% and 0.7\% of difference in AUC and log-loss, respectively. Another aspect worth mentioning is that Higgs Boson data contains quite a few highly correlated variables. The results show that LIFE algorithm outperforms all the rest of models, which indicates LIFE really does an excellent job in predicting on highly correlated structures.

\begin{table}
\centering 
\scalebox{0.85}{
  \begin{threeparttable}
\caption{ResNet18 vs. Its Potential Improvers} 
\begin{tabular}{l c c c c c  } 
\hline\hline 
 Data & Metric & ResNet18 & ResNet18\textunderscore LIFE & ResNet18\textunderscore Xgboost & ResNet18\textunderscore Adam
\\ [0.5ex]
\hline 
                                       &                                          &  0.934    &  \textcolor{red}{\textbf{0.947}}   &   0.944    &  0.935   \\  [-1.5ex]
                                       &\raisebox{1.5ex}{$AUC$}     & (0.030)  & (0.026) & (0.027)  & (0.034) \\  
\raisebox{1.5ex}{MNIST}  &                                          &  0.261    & 0.202   &  \textcolor{red}{\textbf{0.174}}     &  0.332   \\  [-1.5ex]
                                       &\raisebox{1.5ex}{$logloss$}  & (0.046)  & (0.046) & (0.045)  & (0.027) \\ 
 [0.5ex]  

\hline\hline 
\end{tabular}
\label{tab:MNIST_ResNet18}
 \end{threeparttable}
}
\end{table}

We have also investigated whether LIFE algorithm can further improve the performance of trained deep neural network on image data. We here use ResNet18 proposed by (He et al. 2016 \cite{he2016deep}) as an example and apply the algorithms on MNIST data \cite{lecun1999mnist}. The detailed information of data preprocessing can be found in Case 8 from the description lists of real data sets \ref{Case8_MNIST}. After training MNIST data using ResNet18, the output of final convolutional layer has been extracted, which has size $8000 \times 512$, and it is also the input of feed forward neural network (FFNN) in the final step of ResNet18. This $8000 \times 512$ data is then treated as the input of LIFE(Adam). Further, we also attach XGBoost, Adam to ResNet18 and compare the results.

\subsubsection{Classification on Image Data}
\begin{figure}[h]
\center
\includegraphics[width=16cm,height=6cm]{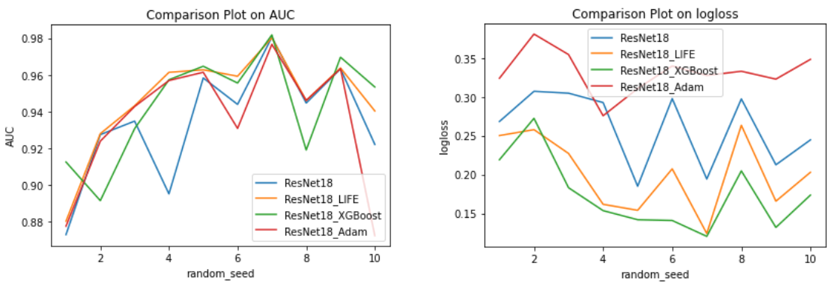}
\caption{ResNet18 vs. Its Potential Improvers}
\label{fig:ResNet18_LIFE}
\end{figure}

Figure \ref{fig:ResNet18_LIFE} shows the performance of LIFE (orange line) is better than ResNet18 (blue line) with consistently larger AUC and smaller log-loss in all of the 10 replications. LIFE is comparable to XGboost in terms of log-loss in general with all 10 values below ResNet18. However, in terms of AUC, XGboost is worse than ResNet18 with significantly lower AUC in two replications (seed=2 and seed=8). On the other hand, Adam (red line) is not able to further improve the performance of ResNet18, which is consistent with what we have discovered in the previous empirical studies. Table \ref{tab:MNIST_ResNet18} also indicates LIFE and Xgboost are both capable of remarkably enhancing a trained deep NN with similar performance. One last discovery is since the input of LIFE contains 512 columns, it also indicates that LIFE can handle high dimensionality quite well in terms of prediction.

\section{Interpretation}\label{sec:int}
Interpretability is the degree to which one human being can understand the cause of a decision or predict the result of a model. The higher the interpretability of a machine learning or deep learning model, the easier it is for someone to comprehend why certain decisions or predictions have been made. A key advantage of LIFE is that it is still an inherently interpretable model. From the perspective of the NN structure, the model is a single-hidden-layer NN with ReLU activation function where all the weights and bias can be easily extracted and visualized. Moreover, the single layer NN with ReLU activation function can be rewritten in the form of local linear model representation, and be interpreted by exploring the patterns of local linear model coefficients. Finally, the main and interaction effects can be identified by exploring and aggregating the local linear coefficients.  

We use the bike sharing data result as an example to illustrate the intrinsic interpretability of LIFE. Bike sharing data is a public dataset hosted on UCI machine learning repository, where there are around $17,000$ observations on hourly (and daily) bike rental counts along with weather and time information between $2011$ and $2012$ in the Capital Bikeshare system. Out of the original $17$ predictors, we removed some non-meaningful and highly correlated ones, leaving us with $9$ predictors to predict hourly rental counts. At the tiny expense of predictive performance, we applied both the base learner selection method shown in Algorithm \ref{algo:bls} in Section \ref{sec:bls} and elastic net to reduce the number of base learners and features so that the final single hidden-layer NN has a small number of signficant neurons and is easier for interpretation.

\begin{figure}[h]
     \centering
   \subfloat[][Neuron Importance]{\includegraphics[width=6cm,height=6.6cm]{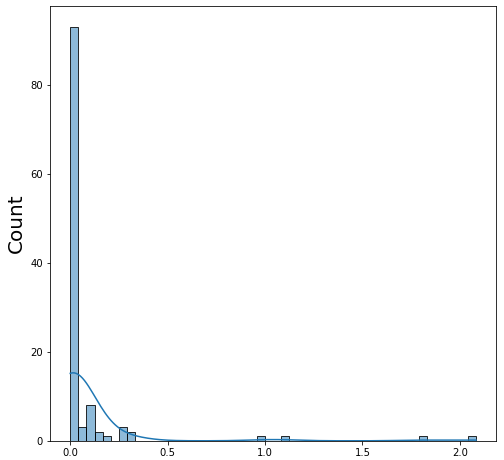}\label{fig:neuron_imp}}
     \subfloat[][Contribution of Variables to Neurons]{\includegraphics[width=11cm,height=7cm]{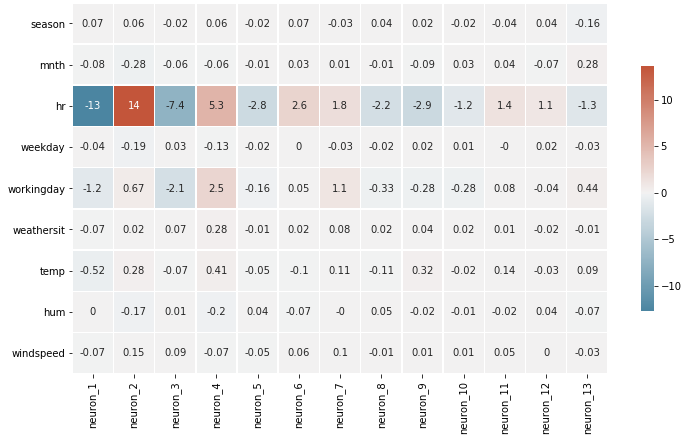}\label{fig:heatmap_imp}} 
     \caption{Variable Importance Detection}
     \label{fig:loss_dec_cla}
\end{figure}

\subsection{Explore the weights and bias of single layer NN}
As LIFE finally generates a single-hidden-layer NN in the third step, we can explore the weights $\hat{w}_k$s and bias $\hat{b}_k$s of the NN directly and identify which variable is important. For bike sharing data, there are finally $116$ new features (or neurons) after base learner selection and elastic net regularization. We measure the neuron importance by  $std(\hat{\beta}_k\sigma(\hat{b}_k+x^T\hat{w}_k))/ std(\hat{f})$, and $\hat{f}$, where std is the standard deviation,  $\hat{f}$  is the predicted value of response variable for regression or log-odds for classification, $\hat{w}_k$s is the neuron weight, and $\hat{b}_k$ is the coefficient for neuron. This quantity measures the importance of neurons/feature by comparing the variation of each feature to the total variance.   The histogram on the neuron importance for the $116$ features in Figure \ref{fig:neuron_imp} shows that there are only $13$ neurons whose importance values are greater than $2\%$ of the maximum importance.

Then we can detect how each variable contributes to each neuron by applying the following measurement:
$$\hat{w}_k\hat{\beta}_k \frac{std(\sigma(\hat{b}_k+x^T\hat{w}_k))}{std(\hat{f})}$$
where we allocate neuron importance to each variable by multiplying $\hat{w}_k$. This contribution measurement can be simply visualized by heatmap between neurons and original variables in Figure  \ref{fig:heatmap_imp} .  It shows that hour(\textit{hr}) and working day(\textit{workingday}) are top significant variables with darker colors for almost each important neuron compared with other variables. Another variable temperature(\textit{temp}) can also be considered to relatively important except \textit{hr} and \textit{workingday}.

\subsection{Treat a single layer NN as a local linear model} \label{local_imp_plot}
As we may have many features in the final wide single layer NN, it is difficult to visualize and explore the weights and bias of all the neurons. Hence, we also propose to interpret single layer NN from local linear model perspective.  Single-layer NN with ReLU function can be considered a type of local linear model. Each linear projection would determine the active or inactive states of the ReLU neurons at hidden layers, which define the layered pattern. The activation region is constructed as a combination of those distinct patterns. Those activation regions are mutually exclusive and regarded as convex polytopes with closed-form boundaries \cite{sudjianto2020unwrapping}. A linear equation can be used in all data points inside the activation region to represent the relationships between response and independent variables. After defining the region each observation belongs to, we can easily extract a linear equation for each region based on estimated weights in the hidden and output layers. The detailed algorithm that performs a linear equation extraction the following:
\begin{algorithm}[H]
\DontPrintSemicolon
  \KwInput{Estimated weights and biases $\{\hat{w}_k,\hat{b}_k\hat{\beta}_k \}_{k=1,\cdots,m_J}$; $\tau$: threshold for the number of observations}
  \KwOutput{Coefficients of local linear equations $\{\hat{E}_t \}_{t=1,\cdots,M}$}
  \KwData{Independent variables $\{x_i \}_{i=1,\cdots,N}$}
Determine if $i^{th}$ observation is active or not in each neuron based on linear projection $\hat{b}_k+x_i^T\hat{w}_k>0$\\
Combine observations into $M$ homogenous activation local regions $R_t, t=1,\cdots,M$, depending on if they have the same set of active neurons ( boundary condition)\\
\For{$t=1,\cdots,M$}   
{ \If{  $\sum_{x_i \in R_t} i>\tau$ }    
{1. Construct activation set $A_t$ of the local region $R_t$ relying on its boundary condition\\
2. Calculate coefficients: $ \hat{E}_t=\sum_{k\in R_t} \hat{w}_k*\hat{\beta}_k$}
}
\caption{Local Linear Equation Extraction}
\label{algo:llm}
\end{algorithm}

We can visualize those linear equations by a parallel coordinate plot, which allows comparing the estimated coefficients of all predictors for different local linear regions. Through the visualization of local linear equations, we can not only have an overview of the importance of each predictor in each region by comparing the magnitude of coefficients, but also check the validity for effect of each predictor on the response variables. It is worth mentioning that those coefficients are comparable after standardizing all the predictors. There are three scenario for a particular independent variable:
\begin{enumerate}
	\item Relatively large coefficients of the variable, compared with others in terms of absolute values and have the same signs, imply that this variable has a significant positive or negative effect on the response variable if all coefficients are positive or negative.
	\item Relatively large coefficients of the variable with both positive and negative signs strongly imply that this variable has inconsistent slopes across local activation regions, which might be due to either its own nonlinear main effect or the interaction effects with other variables.
	\item Small and close-to-zero coefficients indicate that this feature is not important to explain the variation of the response variable and can be removed from the model.
\end{enumerate}

\begin{figure}
     \centering
     \includegraphics[width=6.5in]{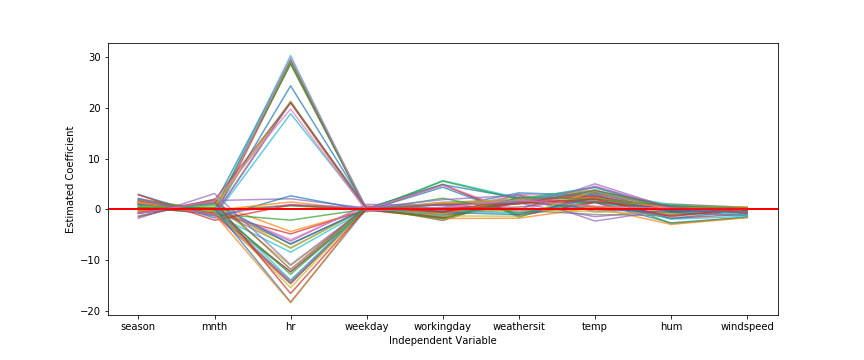}
     \caption{Parallel Corrdiantes Plot for Bikesharing Data}
     \label{fig:local_imp}
\end{figure}

Furthermore, we were able verify if the sign of estimated coefficients of predictor in all regions is consistent with domain knowledge or business sense. Figure \ref{fig:local_imp} displays the estimated coefficients of all predictors in the local activation regions for bike sharing dataset. There are $116$ neurons extracted from NN base learners and $47$ local regions created by Algorithm \ref{algo:llm} and each local region has at least three data points. 
It clearly indicates that hour(\textit{hr}), working day(\textit{workingday}) and temperature(\textit{temp}) are the three most important predictors with relatively higher absolute values of their corresponding coefficients in several local regions, which is pretty consistent with result from Figure \ref{fig:heatmap_imp}. Their estimated coefficients present different directions across local activation regions, which is consistent with our second scenario.

This gives us a hint of interactions between those variables Other variables such as humidity(\textit{hum}) and wind speed(\textit{windspeed}) are insignificant based on their absolute values of estimation coefficients from the plot. Sometimes there are too many local regions and (Sudjianto et al.(2020))\cite{sudjianto2020unwrapping} provides two approaches to simplify and reduce the number of local linear equations—merging and flattening in their paper, where a variety of other diagnostic tools and plots for local linear model have also been provided. 

\subsection{Main and Interaction Effect Detection}
Even though the parallel coordinates plot provides a guideline about the variable importance in each local region, we still need a solid technique to detect nonlinear main effects and interaction effects. To achieve this purpose, we can treat single-hidden-layer NN as a varying coefficient model through linear equation extraction shown in Algorithm \ref{algo:llm}.  As all the local linear equation coefficients are varying over local regions, and region definition depends on predictors, so the coefficients can be treated as a function of predictors in equation \ref{eq:vc_model}.
\begin{equation} \label{eq:vc_model}
\begin{gathered}
\hat{f}_i=\alpha_{0i}+\alpha_{1i}x_{1i}+\cdots+\alpha_{mi}x_{mi}+\cdots+\alpha_{pi}x_{pi},  \ \ \ i=1,\cdots,n, \\
\text{where } \alpha_{mi}=g(x_{1i},\cdots,x_{pi}), 
\end{gathered}
\end{equation}
where $p$ is the number of predictors and $n$  is number of observations. $\hat{f}_i$ is predicted value for regression and predicted log-odds for classification. $\alpha_{mi}$ is the coefficient for $m^{th}$ variable at  $i^{th}$ observation, and could also be a function of all predictors, varying by different observations. Our goal is to investigate what the functional forms of the estimated coefficients are. Therefore, we separate $\alpha_{mi}$ into two components representing main and interaction effects in equation \ref{eq:coef_eff}:
\begin{equation} \label{eq:coef_eff}
\begin{gathered}
\alpha_{mi}=g(x_{1i},\cdots,x_{pi}) = \underbrace{g_{main}(x_{mi})}_\text{main effect}+\underbrace{g_{int}(x_{1i},\cdots,x_{pi})}_\text{interaction effect}.\\
\end{gathered}
\end{equation}
The first term in the equation \ref{eq:coef_eff}  is a function of  $x_{mi}$ , including the intercept of $\alpha_{mi}$, and this term captures the main effect of $x_{mi}$. If $\alpha_{mi}$ has a significant intercept, then linear main effect can be identified; while a strong relationship with $x_{mi}$ indicates a nonlinear main effect. The remaining second term is the function of other predictors and it may or may not contain  $x_{mi}$. This term can be used to detect interactions between $x_{mi}$ and other predictors.  For an illustration, let us look at a simple example with all estimated coefficients constant except $\alpha_{1i}=\theta_0+\theta_1 x_{1i}+\theta_2 x_{2i}$, then the varying coefficient model can be expressed as follows:
\begin{equation} \label{eq:coef_gam_case}
\begin{gathered}
y_i=\alpha_{0i}+\theta_0 x_{1i}+\theta_1 x_{1i}^2+\theta_2(x_{2i})x_{1i}+\alpha_{2}x_{2i}+\cdots+\alpha_{p}x_{pi}+\epsilon_i. 
\end{gathered}
\end{equation}
where $g_{main}(x_{1i})=\theta_0+\theta_1x_{1i}$ and $g_{int}(x_{2i})=\theta_2x_{2i}$. We can easily identify the interaction term between $x_{1i}$ and $x_{2i}$, and $x_{1i}$ shows a nonlinear main effect via its quadratic term. To detect main effects and interaction effects from $\alpha_{mi}$,  we propose the two-stage process below:
\begin{enumerate}
	\item  Check nonlinearity: Calculate conditional expectation $E(\hat{\alpha}_{mi}|x_{mi})$ by smoothing estimated coefficients of predictors against itself \footnote{The smoothing spline is used for illustration and other estimation methods can also be applied.}.  \\ $\hat{\alpha}_{mi} \sim g_{main}(x_{mi})\ \ \ m=1,\cdots,p$.
         \item Check interactions: Remove main effect from $\alpha_{mi}$, and calculate conditional expectation $E(\hat{\alpha}_{mi}-\hat{g}_{main}(x_{mi})|x_{ki})$ by smoothing estimated coefficients of predictors against each other variable.\\
       $\hat{\alpha}_{mi}-\hat{g}_{main}(x_{mi}) \sim g_{k}^m(x_{ki})\ \ \ \ k\neq m$
    \end{enumerate}
\begin{figure}[h]
     \centering
     \includegraphics[width=6in]{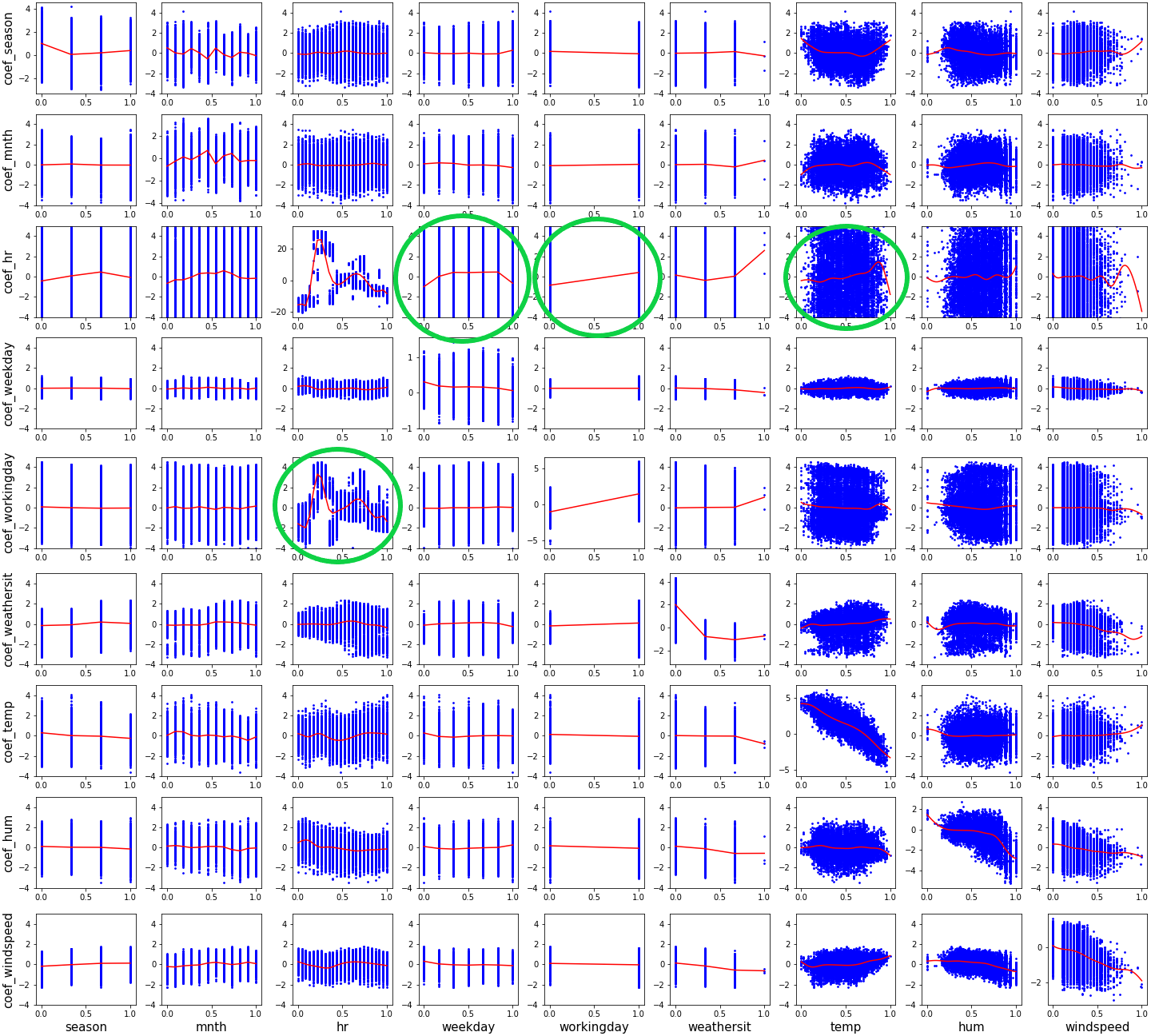}
     \caption{Plot Matrix Between $\alpha_m$ and $x_m$ }
     \label{fig:int_dect}
\end{figure}

We choose to use a two-stage process instead of a one-stage process, as we can estimate its main and interaction effect more accurately in the correlated predictor case and split two effects effectively.  Note that some special interaction effects may not be identified by one-stage process such as $y=\alpha_0+\alpha_1 x_1$ as an example, where $\alpha_1=x_1 x_2$.  In this case, $g_2^1 (x_{2i})$ is zero curve. Fortunately, most common interaction patterns can be identified by our two stage process. As long as $g_k^m (x_{ki})$ has a significant pattern on $x_{ki}$, an interaction effect can be identified. 

\begin{figure}[h]
     \centering
     \includegraphics[width=4in]{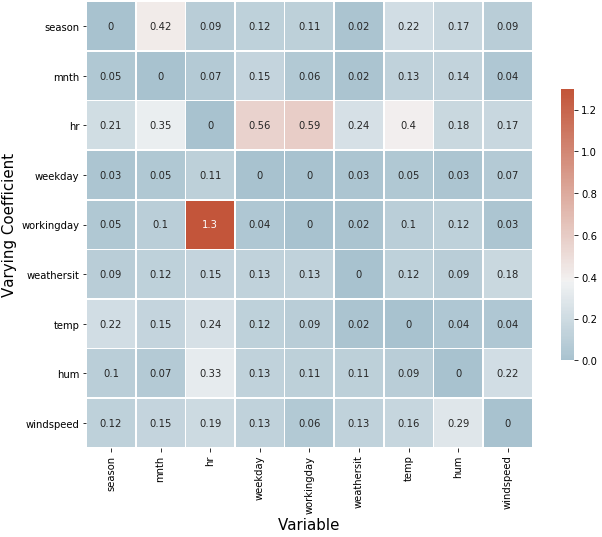}
     \caption{Heatmap for Interaction Measures}
     \label{fig:int_dect_heatmap}
\end{figure}

For the case of bike sharing data, we visualized all pairs of varying coefficients and variables $(\alpha_{mi}\  vs\ x_{ki})$ with scatter plots in Figure \ref{fig:int_dect}. Due to $\frac{\partial \hat{f}(\bf{x})}{\partial x_m} = \alpha_m$ ,  this is also scattered partial derivative plot for $\hat{f}(\bf{x})$ . On top of the scatter plot, we also draw $g_{main} (x_{mi})$ against $x_{mi}$  in the diagonal plots show and $g_k^m (x_{ki})$ agaist $x_{ki}$ in the $(m, k)$ off-diagonal plots .  To further quantify the magnitude of the interaction effects, we calculated weighted standard deviation of $g_{main} (x_{mi})$ and $g_k^m (x_{ki})$ with population density as weight. The heatmap of the interaction measures for bike sharing data is provided in Figure \ref{fig:int_dect_heatmap} where diagonals are masked by zero.

\begin{figure}[h]
     \centering
     \includegraphics[width=5in]{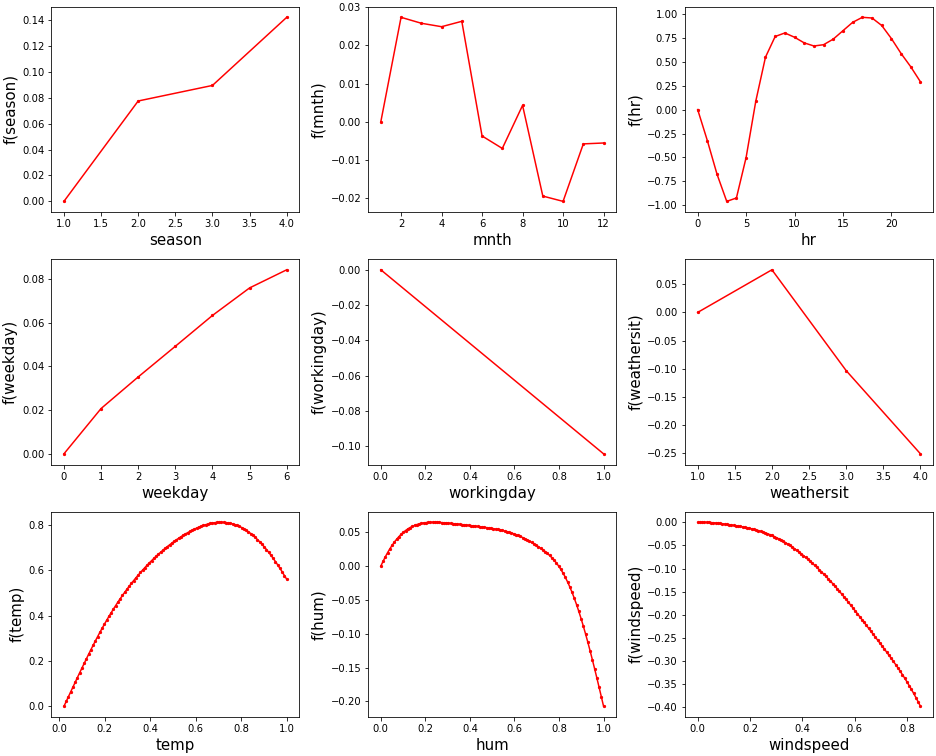}
     \caption{ALE Plots for Predictors Based on LIFE}
     \label{fig:gam_plot2}
\end{figure}
The nonlinear patterns of variables can be clearly spotted in the diagonal plots in Figure \ref{fig:int_dect}. The most important variable hour(\textit{hr}) displays the drastic fluctuation compared with others, indicating its nonlinear effect on response. As evidenced in both Figure \ref{fig:int_dect} and Figure \ref{fig:int_dect_heatmap}, the top three interaction pairs including \textit{hr} vs \textit{workingday}, \textit{hr} vs \textit{weekday} and \textit{hr} vs \textit{temp} can be easily identified.

In addition to interaction detection, we can obtain and visualize the main effect of each predictor directly by aggregating the local linear coefficients.  Due to $\frac{\partial \hat{f}(\bf{x})}{\partial x_m} = \alpha_m$ in the varying coefficient setting, we compute the exact main effect of $x_m$ by constructing a relationship between $f(x_j)$ and $x_j$ based on formula $\int_0^x E(\frac{\partial \hat{f}(\bf{x})}{\partial x_m}|x_m)dx_m$  from Accumulated Local Effects (ALE) plot, where the variable is transformed back to original scale as seen in Figure \ref{fig:gam_plot2}. 
This ALE formulation can be simplified as $\int_0^x E(\alpha_m|x_m)dx_m=\int_0^x g_{main}(x_m)dx_m$ and its numerical implementation of ALE is achieved by the Midpoint Rule.
The main effect for\textit{hr} has two peaks and one trough, which is similar to partial dependence plot from other machine learning algorithms, while the main effect of \textit{temp} and \textit{hum} show a quadratic relationship. More specifically, the peak of bike rentals happen around 7 am and 5-6 pm, while very few people will rent bikes around 3-4 am. People usually prefer to rent bikes in a nice day with moderate temperature and humidity. Both of them are pretty consistent with common sense.

\section{Discussion}\label{dis}
\subsection{Different Sampling Schemes} \label{sec:dss}
The LIFE algorithm can be considered a general framework with three steps, as discussed in the methodology section. LIFE is very flexible and allows users to try different combinations of three steps. The first step presents several data sampling options. In the paper, we use linear projection inside NN neurons to split data and select data points from active region for base learner training, as shown in Figure \ref{fig:tree}.
Those linear projections are obtained with trained NNs in a supervised setting. Given a fixed hyper-parameter setup for LIFE, we have implemented our method with different sampling choices including NN projection, Random projection, Bootstrapping. In Table  \ref{tab:data_sample}, we can easily see that all the ensemble methods outperform a single single-hidden-layer NN model optimized by LLA or Adam. Most importantly, sampling by NN linear projection is better than other sampling methods that create subsets in a random way.
\begin{table}[h]
\centering 
\scalebox{0.85}{
  \begin{threeparttable}
\caption{Sampling Method Comparison} 
\begin{tabular}{l c c c c | c} 
\hline\hline 
 Model & Metric & NN projection & Random projection & Bootstrapping & LLA or Adam
\\ [0.5ex]
\hline 
&  &\textcolor{red}{\textbf{1.060}} & 1.095 & 1.138 & 1.110\\  [-1.5ex]
 &\raisebox{1.5ex}{$RMSE$}  &(0.017) & (0.023) & (0.034) & (0.065)\\  
MIM (Regression)   &  &\textcolor{red}{\textbf{0.965}} & 0.962 & 0.960  & 0.961\\[-1.5ex]
 &\raisebox{1.5ex}{$R^2$} & (0.001) & (0.002) & (0.003) & (0.005)\\ 
\hline
  &  &\textcolor{red}{\textbf{0.500}} & 0.525 & 0.519 & 0.527 \\  [-1.5ex]
 &\raisebox{1.5ex}{$RMSE$}  &(0.012) & (0.020) & (0.018) & (0.011) \\  
California Housing   &  &\textcolor{red}{\textbf{0.812}} & 0.792 & 0.797 & 0.791\\[-1.5ex]
 &\raisebox{1.5ex}{$R^2$}  & (0.009) & (0.016) & (0.014)  & (0.008)\\ 

\hline
  &  &\textcolor{red}{\textbf{0.940}} & 0.935 & 0.936  & 0.920\\   [-1.5ex]
 &\raisebox{1.5ex}{$AUC$}  &(0.004) & (0.004) & (0.003) & (0.003)\\  
MIM (Classification)   &  &\textcolor{red}{\textbf{0.268}} & 0.277 & 0.275 & 0.421\\[-1.5ex]
 &\raisebox{1.5ex}{$logloss$}  & (0.012) & (0.009) & (0.007) & (0.075)\\ 
\hline
  &  &\textcolor{red}{\textbf{0.938}} & 0.929 & 0.927 & 0.909\\  [-1.5ex]
 &\raisebox{1.5ex}{$AUC$}  &(0.002) & (0.003) & (0.003) & (0.008)\\  
Gamma Telescope   &  &\textcolor{red}{\textbf{0.288}} & 0.307 & 0.314 &  0.371\\[-1.5ex]
 &\raisebox{1.5ex}{$logloss$}  & (0.007) & (0.007) & (0.007) & (0.019)\\ 

\hline\hline 
\end{tabular}
 \begin{tablenotes}
      \small
      \item Note: The different sampling methods in the first step of the general framework of LIFE are tested on two simulated data and two real data. NN projection performs data partition by linear projection inside neurons of NNs and then samples data from the selected region, which is the main part of LIFE algorithm used in the paper.  Random projection performs data partition by random linear projection, where weights and biases are randomly drawn from standard normal. The bootstrapping selects observations randomly from training set. The number colored in red represents optimal result for this metric. LLA is used to optimize single-hidden-layer NN base learner in the regression case, which Adam is used for classification case. 
    \end{tablenotes}
\label{tab:data_sample}
 \end{threeparttable}
}
\end{table}

\subsection{Base Learner Selection}  \label{sec:bls}
For model aggregation and pruning, we can prune neurons to have a single layer NN with fewer neurons. We used Elastic net for pruning due to its simplicity, but pruning methods besides Elastic net can also be considered. Based on properties of LIFE, we also developed an alternative pruning method called base learner selection to reduce the number of nodes in the final step. It is assumed that LIFE works well because the correlations of prediction errors are not strong among different base learners. Therefore, we can remove base learner one by one, according to the correlation between its prediction errors and prediction errors from other base learners. In this way, we can still maintain diversity and solve overfitting issue by keeping fewer necessary base learners without sacrificing predictive performance a lot. This method is thoroughly described by Algorithm \ref{algo:bls}.
\begin{algorithm}[H]
\DontPrintSemicolon
  \KwInput{$B_j$: trained base learner $j=1,\cdots,M$; $\tau$: threshold}
  \KwOutput{$B_j$: selected base learner $j\in\{1,\cdots,M\}$}
  \KwData{Training set $\{x_i,y_i\}_{i=1,\cdots,N}$}
      \For{$j=1,\cdots,M$} {
     1. $\hat{y}_i^{(j)} = B_j(x_i)$ \\
     2. $r_i^{(j)}=y_i-\hat{y}_i^{(j)}$
 }   
      \For{$j=1,\cdots,M$} {
     1. Train a linear model $r_i^{(j)} \sim c_0 + c_1r_i^{(1)}+\cdots +c_{j-1}r_i^{(j-1)}+ c_{j+1}r_i^{(j+1)}+\cdots + c_{M}r_i^{(M)}$\\
     2. Collect $R^2_j$ from this linear model
 }   
Remove $B_j$ based on value of $R^2_j$ in descending order until $\tau$ is achieved (the one with highest $R^2$ will be removed first).
\caption{Base learner selection}
\label{algo:bls}
\end{algorithm}

\begin{figure}[h]
     \centering
    \subfloat[][GAM]{\includegraphics[width=3.2in]{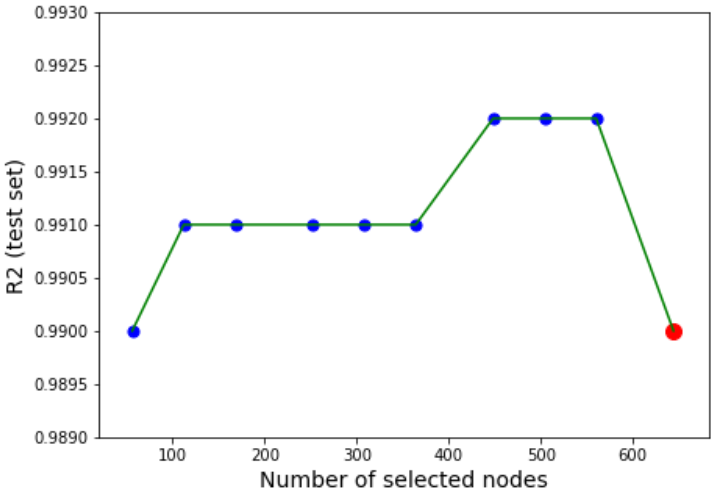}\label{fig:gam_base}}
     \subfloat[][MIM]{\includegraphics[width=3.2in]{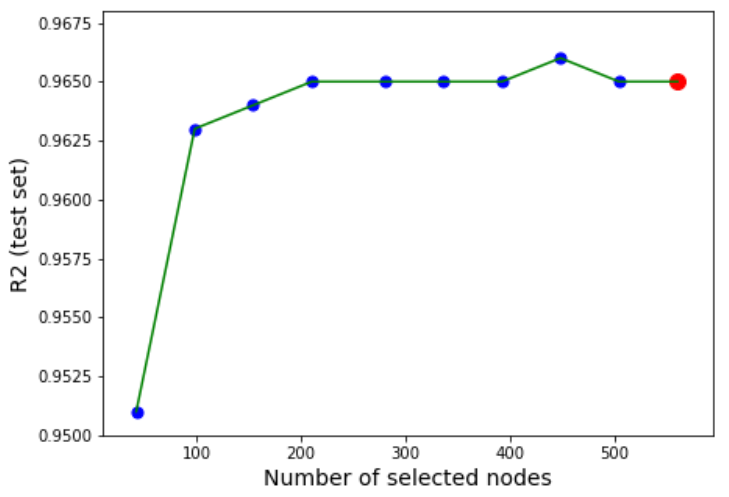}\label{fig:mim_base}}
     \caption{Relationship between $R^2$ and number of neurons. The red point indicates one using linear regression as a final step without base learner selection, while blue points indicates model aggregation by base learner selection with different threshold.}
     \label{fig:base}
\end{figure}

In Algorithm \ref{algo:bls}, the threshold $\theta$ is the percentage of base learners you want to retrieved from a pool of candidates. When there is a large number of neurons in LIFE setting, the elastic-net is usually computational expensive and the base learner selection is a good alternative by parallel computation. Moreover, we can combine these two pruning methods to achieve a simpler NN model from wide NN faster. We also illustrate it using two simulated model (GAM and MIM). The plots  (a) and (b) in Figure \ref{fig:base} show the relationship between $R^2$ and number of hidden neurons 
for the feature extraction. Setting different thresholds in the base learner selection Algorithm \ref{algo:bls}, we can construct single-hidden-layer NN with different number of neurons. In general, base learner selection algorithms can effectively reduce number of neurons to produce a simpler model without sacrificing predictive performance or obtain even better results.

\section{Conclusion}\label{sec:con}
In this paper, we have proposed a novel algorithm that fits single-hidden-layer NN to achieve three goals: ensuring competitive predictive performance, boosting computational efficiency, and preserving the interpretability of the model. Unlike traditional NN training methods, we train it in an iterative way through multiple NNs layer-by-layer training and then effectively combine them via neural nodes flattening. We have evaluated the performance of our approach using simulated and empirical data in terms of predictive accuracy and computational efficiency and found that it consistently outperforms single-hidden-layer NN trained directly by LLA or Adam optimizer and achieves competitive results as those of Xgboost. 

This superior performance lies in three reasons: First, as an ensemble method, the LIFE algorithm performs data sampling through linear projection inside neural nodes, which creates diversity among the models and contributes to bias and variance reduction of prediction from combined models. Second, the LIFE algorithm takes advantage of single-hidden-layer NN structure to combine multiple narrow single-hidden-layer NNs into a wide one via neural nodes flattening Third, LIFE algorithm benefits from leveraging parallel computing to train multiple NNs on subsets of data simultaneously. Moreover, the base learner selection method is introduced in the paper to help us prune redundant neural nodes and produce a more parsimonious model after several iterations of the LIFE algorithm. We have also proposed a new method for main and interaction detection from the perspective of interpretation.

\bibliographystyle{abbrvnat}
\bibliography{LIFE}

\section*{Appendix}
\subsection*{Loss Decomposition in the Log-odds Space}\label{sect.log}
The cross-entropy loss for one observation in the log-odds space can be written as:
\begin{equation} \label{eq:loss_cal}
\begin{gathered}
l(y_i,f_i)=y_ilog(1+e^{-f_i})+(1-y_i)log(1+e^{f_i}) 
\end{gathered}
\end{equation}
Therefore, the average cross entropy loss is expressed as:
\begin{equation} \label{eq:cal_dec2}
\begin{gathered}
\sum_{i=1}^N [y_ilog(1+e^{-f_i})+(1-y_i)log(1+e^{f_i})] \\
=\underbrace{\sum_{i=1}^N\sum_{j=1}^M \beta_j[y_ilog(1+e^{-f_i^{(j)}})+(1-y_i)log(1+e^{f_i^{(j)}})]}_\text{accuracy} \\
-\underbrace{\sum_{i=1}^N\sum_{j=1}^M \beta_j \{\frac{e^{f_i^{(j)\star}}+e^{-f_i^{(j)\star}}+2}{[(1+e^{-f_i^{(j)\star}})(1+e^{f_i^{(j)\star}})]^2}\}(f_i- f_{i}^{(j)})^2}_\text{diversity}
\end{gathered}
\end{equation}

\subsection*{Local Linear Approximation (LLA) Algorithm}\label{sect.lla}
The algorithm is extracted from Zeng, et al (2020) \cite{Zeng2020LLA}. 
Note that $h^{(0)}=x$. Then one hidden-layer FNN with $J_1$ nodes can be expressed as follows
$ h^{(1)}=m{\sigma}(Wh^{(0)} + b)=(\sigma(w^T_1x + b_1),
\cdots, \sigma(w^T_{J_1}x + b_{J_1}))^T,
$
where we drop the subscripts of $W$ and $b$ for ease of presentation. We set  $\sigma(z)=\max\{z,0\}$, the ReLU activation function.
To estimate the weights and biases, we minimize a nonlinear LS function
$$ l_1(\theta)=\sum_{i=1}^n \{y_i-\beta_0-\sum_{j=1}^{J_1} \beta_j \sigma(w_j^Tx_i+b_j)\}^2 $$
Given $w_j^{(c)}$ and $b_j^{(c)}$ in the current step, we propose to approximate $\sigma(x^Tw_j + b_j)$ by a linear function based on the first-order Taylor expansion of $\sigma(z)$:
$$
\sigma(x^Tw_j + b_j) \approx \sigma(x^Tw_j^{(c)} + b_j^{(c)}) +
\{(x^Tw_j+ b_j)- (x^Tw_j^{(c)}+b_j^{(c)})\}I(x^Tw_j^{(c)}
+ b_j^{(c)}>0) $$
for $j=1,\cdots, J_1$. Thus,
$$
\beta_j\sigma(x^Tw_j + b_j) \approx \beta_j\sigma(x^Tw_j^{(c)} + b_j^{(c)}) +
\gamma_j I(x^Tw_j^{(c)} + b_j^{(c)}>0) + m{\eta}_j^Tx I(x^Tw_j^{(c)}
+ b_j^{(c)}>0) $$
where $\gamma_j=\beta_j(b_j-b_j^{(c)})$ and  $m{\eta}_j= \beta_j(w_j-w_j^{(c)})$.
Define $z_{1ij}=\sigma(x_i^Tw_j^{(c)}+ b_j^{(c)})$, $z_{2ij}=I(x_i^Tw_j^{(c)} + b_j^{(c)}>0),
$ and $z_{3ij}=x_i I(x_i^Tw_j^{(c)}+ b_j^{(c)}>0)\},$ $j=1\cdots, J_1$. Further define
$z_{1i}=[z_{1i1},\cdots,z_{1iJ_1}]$, $z_{2i}=[z_{2i1},\cdots,z_{2iJ_1}],$
$z_{3i}=[z_{3i1}^T,\cdots,z_{3iJ_1}^T],$ and $z_i=[z_{1i}, z_{2i},z_{3i}]^T$, which is
a $J_1(p+2)$-dimensional vector.  the objective function is approximated by
$$
\sum_{i=1}^n \{y_i-\beta_0-\sum_{j=1}^{J_1}\{\beta_jz_{1ij} +
\gamma_j z_{2ij} + m{\eta}_j^Tz_{3ij}\}\}^2,
$$
which is the LS function of linear regression with the response $y_i$
and predictors $z_i$. Denote the resulting LS estimate of $\beta_j$, $\gamma_j$ and
$m{\eta}_j$ by $\hat{\beta}_j$, $\hat{\gamma}_j$ and $\hat{m{\eta}}_j$, respectively.
By the definition of $\gamma_j$ and $m{\eta}_j$, we can update $b_j$ and $w_j$ as shown in the step 2 of algorithm
If $|\hat{\beta}_j|$ is very close to zero, one may simply set $b_j^{(c+1)}=b_j^{(c)}$
and $w_j^{(c+1)}=w_j^{(c)}$. Thus, we may estimate $W$ and $b$ by iteratively
 and regressing $y_i$ on the updated $z_i$. The procedure can be summarized as the following algorithm.
\begin{enumerate}
\item Set initial value for $W^{(0)}=[w_1^{(0)},\cdots,w_{J_1}^{(0)}]^T$ and
$b_j^{(0)}$, and let $c=0$.
\item Calculate $z_i$ defined in the text based on $w_j^{(c)}$ and $b_j^{(c)}$, obtain the least squares estimate (LSE) $\hat{\beta}_j$s, $\hat{\gamma}_j$s and
$\hat{\eta}_j$s by running a linear regression $y_i$ on covariate $z_i$, and update the biases and
weights by
$$ b_j^{(c+1)}=b_j^{(c)} + \hat{\gamma}_j/\hat{\beta}_j,
\ \text{and}\  w_j^{(c+1)}=w_j^{(c)} + \hat{\eta}_j/\hat{\beta}_j. $$
if $|\hat{\beta}_j|\ge\varepsilon$, where $\varepsilon$ is a constant for numerical stability
and is set to be $10^{-3}$ in our numerical experiment, and keep the corresponding biases and weights unchanged
if $|\hat{\beta}_j|<\varepsilon$.
\item Set $c=1,2,\cdots, $ and repeat Step 2 until the criterion of algorithm convergence meets.
\end{enumerate}

\subsection*{Functional Forms for Classification Case in Simulation Study} \label{sect.sim_model}
In classification case, we use three function forms including Generalized Additive Model (GAM), Additive Index Model (AIM) and Multiple Index Model (MIM) expressed as follows, where  $\epsilon_i  \sim N(0,1), i=1,\dots,N$.
\begin{equation} \label{eq:sim_gam_binary}
\begin{gathered}
GAM:   f( \boldsymbol{x}_i) = 1.5 x_{1i} + 4\sqrt{| -2.5 x_{2i}|} +2 |x_{3i}|+ 4 \exp( -\frac{3}{14} x_{4i})+ 4 \log( 1.5 |x_{5i}|) \\
\hspace{-5cm} - 4 \max(1,x_{6i})+\epsilon_i. 
\end{gathered}
\end{equation}
\begin{equation} \label{eq:sim_aim_binary}
\begin{gathered}
AIM:   f( \boldsymbol{x}_i) = \log(|   3 x_{1i} - 2.5 x_{2i}+ 2 x_{3i} - 1.5 x_{4i} |)+\exp \{   ( - 1.5 x_{4i} + 1.5 x_{5i} -  x_{6i} ) /11 \}+ \epsilon_i.
\end{gathered}
\end{equation}
\begin{equation} \label{eq:sim_mim_binary}
\begin{gathered}
MIM:   f( \boldsymbol{x}_i) = \exp(0.03 x_{1i} - 0.025 x_{2i}) x_{3i} - \frac{ 3 x_{4i}}{1+ 1.5 |x_{5i}|}+ 2\max(1, x_{6i})+\epsilon_i.
\end{gathered}
\end{equation}

\subsection*{Real Datasets} \label{sect.data}
\subsubsection*{Regression}
\textbf{Case 1: Abalone.}
Original data comes from a Marine Resources Division Marine Research Laboratories in Australia. The goal is to predict the age of abalone from physical measurements, which
is determined by cutting the shell through the cone, staining it, and counting the number of rings through a microscope. From the original data examples with missing values were removed, and the ranges of the continuous values have been scaled for use.
Therefore, there are $4177$ observations left and $9$ variables including an integer response variable  (number of rings),  one categorical predictor (sex) and seven continuous predictors (Length, Diameter, Height, Whole weight, Shucked weight, Viscera weight, Shell weight).

\textbf{Case 2: Airfoil.}
It is NASA data set, obtained from a series of aerodynamic and acoustic tests of two and three-dimensional airfoil blade sections conducted in an anechoic wind tunnel.
It comprises different size NACA 0012 airfoils at various wind tunnel speeds and angles of attack. There are $1503$ observations and $6$ variables, where the response variable is scaled sound pressure level in decibels and there are five independent variables containing frequency in Hertzs,  angle of attack in degrees, chord length in meters, free-stream velocity, in meters per second, suction side displacement thickness in meters.

\textbf{Case 3: Aquatic Toxicity.}
This dataset was used to develop quantitative regression QSAR models to predict acute aquatic toxicity towards the fish Pimephales promelas (fathead minnow) on a set of 908 chemicals. to predict acute aquatic toxicity towards Daphnia Magna as a response variable called  LC50. Without missing values, it contains values for 8 predictors (molecular descriptors) of 546 chemicals (TPSA, SAacc, H-050, MLOGP, RDCHI, GATS1p, nN, C-040). 

\textbf{Case 4: Bike Sharing.}
This dataset contains the hourly and daily count of rental bikes between years 2011 and 2012 in Capital bike share system with the corresponding weather and seasonal information.
Specifically, the bike sharing dataset contains $17379$ data points of $16$ predictors. Out of the $16$ independent variables, we removed two non-meaningful information as well as two response-related information. This leaves us with $12$ variables including hour, temperature, feeling temperature, humidity, wind speed, season, workingday
weekday and weather situation, year and month.

\textbf{Case 5: California Housing.}
This dataset was derived from the 1990 U.S. census, using one row per census block group and the target variable is the median house value for California districts. A block group is the smallest geographical unit for which the U.S. Census Bureau publishes sample data (a block group typically has a population of 600 to 3000 people).
It consists of 20640 observations and the independent variables include longitude, latitude, housing median age, medium income, population, total rooms, total bedrooms and households.

\textbf{Case 6: CASP.}
This is a dataset of Physicochemical Properties of Protein Tertiary Structure, which is taken from CASP 5-9. The goal is to predict RMSD-size of residue given other physical attributes. There are 45730 observations and $9$ predictors including total surface area, non-polar exposed area, fractional area of exposed non polar residue, fractional area of exposed non polar part of residue, molecular mass weighted exposed area, average deviation from standard exposed area of residue, Euclidian distance, secondary structure penalty, special Distribution constraints (N,K Value).

\textbf{Case 7: Electrical Grid.}
The local stability analysis of the 4-node star system (electricity producer is in the center) implementing Decentral Smart Grid Control concept. There are $10000$ observations and 11 predictors including $tau[x], x=1,2,3,4$ which are reaction time of participant, $p[x],x=2,3,4$ which are nominal power consumed (negative) divided by produced(positive)(real), and $g[x], x=1,2,3,4$ that are coefficient (gamma) proportional to price elasticity. The continuous response variable is  the maximal real part of the characteristic equation root.

\subsubsection*{Classification}
\textbf{Case 1: Bank Marketing.}
The data provides information regarding direct marketing campaigns of a Portuguese banking institution. The classification goal is to predict if the clients, who were contacted based on at least two phone calls in general, would like to subscribe a term deposit or not. The data contains 45211 examples and 16 variables including 6 numerical variables (age, balance, day, campaign, pdays, previous) and 9 categorical variables (job, marital, education, default, housing, loan, contact, month, poutcome). Notice that input variable `duration' is not included in the analyses based on the suggestions from the provider of this dataset since it highly affects the output target and thus the variable should be discarded due to the intention of building a realistic predictive model. 

\textbf{Case 2: Breast Cancer Wisconsin.}
Data comes from original Wisconsin Breast Cancer Database, with purpose of detecting if the tissue is benign or malignant. The original dataset contains 699 instances and 9 attributes. After deleting rows with missing values, finally 683 instances are used in our analyses. All predictive variables are numerical with a scale of 1 to 10, including Clump Thickness, Uniformity of Cell Size, Uniformity of Cell Shape, Marginal Adhesion, Single Epithelial Cell Size, Bare Nuclei, Bland Chromatin, Normal Nucleoli and Mitoses.

\textbf{Case 3: Higgs Boson.}
This dataset was built from official ATLAS full-detector simulation in 2014 that mixes `Higgs to tautau' events with different backgrounds, where the events in which Higgs bosons were produced are comprised in the signal sample, while other known processes mimicking the signal are considered as background noise. The dataset contains 818238 events and 63 variables, among which a few are highly correlated. The objective is to detect signal from background based on characteristics of events such as mass (estimated, transverse or invariant), transverse momentum, centrality of the pseudo rapidity of the lepton, azimuth angle, etc.

\textbf{Case 4: Home Lending.}
This is an internal First Lien modeling data from Wells Fargo, which was constructed from the entire panel dataset of loans provided from the Consumer Lending Group's (CLG) data modeling team. The goal is to predict the probability of troubled loans based on fico score, unemployment rate, delinquency status, loan to value ratio (LTV), total employment, etc. The toy dataset used in this paper contains 1 million observations and 43 variables, which was selected from 210 million observations obtained by stacking. Training and testing data are split based on `account id' to prevent overfitting. 

\textbf{Case 5: MAGIC Gamma Telescope.}
The dataset comes from Major Atmospheric Gamma Imaging Cherenkov Telescope project (MAGIC), which is Monte-Carlo generated to simulate registration of high energy gamma particles in a ground-based atmospheric Cherenkov gamma telescope using the imaging technique. The objective is to discriminate the images caused by signals (primary gammas) from the images caused by background (hadronic showers initiated by cosmic rays) based on some characteristics such as the attributes of ellipse (fLength, fWidth, fM3Long, fM3Trans, fAlpha, fDist) and attributes on pixels (fSize, fConc, fConc1, fAsym). The dataset contains 19020 observations and 10 continuous predictive variables.

\textbf{Case 6: Mushroom.}
This dataset includes descriptions in terms of physical characteristics of samples related with 23 species of gilled mushrooms. Each species is identified as poisonous or edible. The purpose is to predict the probability of a species being poisonous given the attributes of the species such as cap (shape, surface, color), odor, gill (attachment, spacing, size, color), stalk (surface and color above or below ring), veil (type, color), ring (number, type), spore print color, population and habitat, all of which are categorical variables. The dataset contains 8124 instances and 21 variables.

\textbf{Case 7: Ringnorm.}
 It is a simulated dataset as an implementation of Leo Breimans ringnorm example with 7400 rows and 21 columns. Two classes contained in the response variable, within each class explanatory variables are drawn from a multivariate normal distribution.

\textbf{Case 8: MNIST Data.} \label{Case8_MNIST}
Data are originally obtained from the MNIST database of handwritten digits collected from hundreds of Census Bureau employees and high-school students. The response variable contains signal digit from $0$ to $9$, which are further centered into $28 \times 28$ gray-scaled images. Then the images are translated into regular dataset with dimension 60,000 by 784, indicating there are 60,000 images and $28 \times 28$ positions of the field. Digit `1' is considered as signal and all the other 9 digits are considered as noise and are re-encoded as `0'. Training, validation and testing data are randomly selected from these 60,000 observations for 10 times with size 8000, 2000, 2000, respectively. Images in the training dataset are then further rotated by 20 degrees. 

\end{document}